\documentclass[journal]{IEEEtran}
\usepackage{amsmath,amssymb,amsfonts,bm}
\usepackage{balance}
\usepackage{algorithmic}
\usepackage{algorithm}
\usepackage{array}
\usepackage[caption=false,font=normalsize,labelfont=sf,textfont=sf]{subfig}
\usepackage{textcomp}
\usepackage{stfloats}
\usepackage{url}
\usepackage{verbatim}
\usepackage{graphicx}
\usepackage{cite}

\usepackage{tikz}
\usepackage{booktabs}
\usepackage{multirow}
\usepackage[hidelinks]{hyperref}

\begin{document}

\title{Can Deep Learning be Applied to Model-Based Multi-Object Tracking?}

\author{Juliano Pinto, Georg Hess, William Ljungbergh, Yuxuan Xia, \textit{Graduate Student Member, IEEE}, \\Henk Wymeersch, \textit{Senior Member, IEEE}, Lennart Svensson, \textit{Senior Member, IEEE}\thanks{The authors are with the Department of Electrical Engineering, Chalmers University of Technology, 41296 Gothenburg, Sweden.\\This work was supported, in part, by a grant from the Chalmers AI Research Centre Consortium. Computational resources were provided by the Swedish National Infrastructure for Computing at C3SE, partially funded by the Swedish Research Council through grant agreement no. 2018-05973.}}



\maketitle

\begin{abstract}
Multi-object tracking (MOT) is the problem of tracking the state of an unknown and time-varying number of objects using noisy measurements, with important applications such as autonomous driving, tracking animal behavior, defense systems, and others. 
In recent years, deep learning (DL) has been increasingly used in MOT for improving tracking performance, but mostly in settings where the measurements are high-dimensional and there are no available models of the measurement likelihood and the object dynamics. 
The model-based setting instead has not attracted as much attention, and it is still unclear if DL methods can outperform traditional model-based Bayesian methods, which are the state of the art (SOTA) in this context.
%
%
In this paper, we propose a Transformer-based DL tracker and evaluate its performance in the model-based setting, comparing it to SOTA model-based Bayesian methods in a variety of different tasks. Our results show that the proposed DL method can match the performance of the model-based methods in simple tasks, while outperforming them when the task gets more complicated, either due to an increase in the data association complexity, or to stronger nonlinearities of the models of the environment.

\end{abstract}


\section{Introduction}
Multi-object tracking (MOT) is the problem concerned with recursively estimating the state of an unknown and time-varying number of objects, based on a sequence of noisy sensor measurements. The objects of interest can enter and leave the field-of-view (FOV) at any time, they do not always generate measurements at every time-step, and there can be false measurements originating from sensor noise and/or clutter. Methods capable of tracking objects under these conditions are required for a diverse set of important applications, including tracking animal behavior \cite{mot_for_tracking_animals}, pedestrian tracking \cite{mot_in_pedestrian_tracking2}, autonomous driving \cite{mot_in_autonomous_driving1}, oceanography \cite{mot_in_oceanography}, military applications \cite{mot_in_military}, and many others. Methods to solve the MOT problem depend on whether they operate in the \emph{model-based} or  \emph{model-free} setting. In the model-based setting, accurate and tractable models of the measurement likelihood as well as the object dynamics are available to the MOT designer. In contrast, under the model-free setting, such models are unavailable or intractable, e.g., due to high-dimensional measurements such as image or video data \cite{mot_in_pedestrian_tracking2, mot_challenge}. 

In recent years, deep learning (DL) has been increasingly applied to the field of model-free MOT, resulting in new breakthroughs to state-of-the-art (SOTA) performance \cite{mot_challenge, deep_learning_video_tracking_survey, deep_learning_tracking_survey}. Many works use DL methods to aid in the solution of some of the subtasks for MOT, such as object detection \cite{fasterrcnn, ssd, yolo}, extracting high-level features from input data such as images \cite{mot_with_appearance_feature}, associating new measurements to existing tracks \cite{chen2017enhancing}, managing track initialization/termination \cite{milan2017online}, and predicting motion models \cite{mot_for_motion_models}, to name a few. 
Others attempt to solve the entire (or almost the entire) MOT task using DL, with architectures based on extensions of object detectors \cite{tracking_without_bells_and_whistles}, convolutional neural networks \cite{dmm-net, deep_affinity_network, tubetk, fairmot, mots}, graph neural networks \cite{gnn3dmot, braso2020learning, graph_network_for_mot} or, more recently, the Transformer \cite{transformer_paper} network \cite{TrackFormer, motr, transmot, transtrack_mot}. 
%

In the model-based setting, filters based on the random finite set (RFS) formalism using multi-object conjugate priors \cite{pmbm_older, GLMB} can provide closed-form Bayes-optimal solutions to MOT and achieve state-of-the-art performance \cite{mahler2007}. Yet, due to the unknown correspondence between objects and measurements, also known as the data association problem, the number of possible associations increases super-exponentially over time, and these methods must therefore resort to heuristics such as pruning/merging for remaining computationally tractable \cite{PMBM, GLMB}. This inevitably leads to a deterioration of tracking performance. Moreover, when the measurement and/or motion models are nonlinear, one must rely on Gaussian approximations or sequential Monte Carlo methods to handle the nonlinearity \cite{sarkka2013bayesian}, which may further impact the tracking performance.

In contrast, DL methods are able to directly learn a mapping from sequences of measurements to state estimates in a data-driven fashion, thus sidestepping the complexity of dealing with data associations explicitly and therefore the need to resort to heuristics for maintaining computational tractability. In specific, Transformer-based models have shown great promise in sequence-to-sequence function approximation on a variety of contexts \cite{alphafold, transformers_for_translation, transformers_in_tts, image_transformer}, including the model-free MOT setting \cite{TrackFormer, transmot, transtrack_mot, SoDA, motr}, and have the potential to outperform traditional model-based trackers for complex tracking problems. However, to the best of our knowledge, no prior work with the exception of our preliminary analysis \cite{mt3} has investigated how well DL methods compare against traditional SOTA model-based Bayesian methods such as \cite{jpda_filter, mht, pmbm_older, GLMB} in such scenarios.

In this paper, we compare the capabilities of DL-based trackers compared to traditional Bayesian filters in the model-based context. Specifically, we propose a novel, high-performing DL method for MOT, based on the Transformer architecture \cite{transformer_paper}: the MultiTarget Tracking Transformer v2 (MT3v2). In contrast to most of the existing DL-based SOTA in model-free MOT, MT3v2 is specifically tailored for the model-based setting with low-dimensional measurements, and we use it as a proof of concept to illustrate the potential of DL in this setting. We perform a comprehensive comparison of its tracking performance against two SOTA Bayesian filters: the Poisson multi-Bernoulli mixture filter (PMBM) \cite{pmbm_older} and the delta-generalized labeled multi-Bernoulli filter ($\delta$-GLMB) \cite{GLMB}. The comparison is done on four different tasks with different data association complexities and measurement model nonlinearities. Our results show that the DL tracker achieves comparable performance to the traditional Bayesian filters in simple tasks, while being able to outperform them in scenarios with higher data association complexity and/or strong model nonlinearities, providing evidence of the applicability of deep-learning trackers also in the model-based setting. Our specific contributions are:
\begin{itemize}
    
    \item A novel, high-performing DL tracker based on the Transformer architecture (MT3v2). The proposed architecture provides uncertainty estimates in addition to state estimates, and uses multiple improvements compared to standard Transformers, including a selection mechanism for providing sample-specific object queries, a decoder that iteratively refines estimates, and a learned temporal encoding of the measurement sequences.
    
    \item An uncertainty-aware loss function formulation suited for training DL-based trackers, together with a contrastive auxiliary loss for improving training speed and final performance.
    
    \item An in-depth performance evaluation and comparison with respect to SOTA Bayesian model-based MOT methods under a realistic radar measurement model with non-linearities and finite FOV.  The evaluation considers a recently proposed uncertainty-aware MOT performance measure \cite{nll_letter} and the standard generalized optimal sub-pattern assignment (GOSPA) metric. 
\end{itemize}

The rest of this paper is organized as follows. Section \ref{sec:problem_formulation} introduces the modeling assumptions for MOT and its problem formulation. This is followed by Section \ref{sec:background} which provides a background on the Transformer architecture used in the DL method MT3v2, which is explained in detail in Section \ref{sec:mt3v2}. Section \ref{sec:evaluation_setting} provides all the details regarding our evaluation protocol: a description of the motion and measurement models used, implementation details about MT3v2 and the benchmarks, and the performance measures used. Lastly, Section \ref{sec:results} describes the results obtained, followed by a conclusion in Section \ref{sec:conclusion}.

\subsubsection*{Notations}
Throughout the paper we use the following notations: Scalars are denoted by lowercase or uppercase letters with no special typesetting ($x$), vectors by boldface lowercase letters ($\mathbf x$), matrices by boldface uppercase letters ($\mathbf X$), and sets by blackboard-bold uppercase letter ($\mathbb X$). Sequences are indicated by adding subscripts or superscripts denoting their ranges to the typesetting that matches their elements (e.g., $\mathbf x_{1:k}$ is a sequence of vectors, $\mathbb X_{1:j}$ of sets), and arrays by adding multiple such ranges (e.g., $\mathbf x_{1:k, 1:n}$). The number of elements in a set $\mathbb X$ is denoted $|\mathbb X|$, and we further define $\mathbb N_a\doteq\{i ~:~ i\leq a~,~ i\in\mathbb N\}$.










\section{Multitarget Model and Problem Formulation}
\label{sec:problem_formulation}
\subsection{Measurement and Transition Model}
For the analysis carried out in this paper, we use the standard multitarget transition and observation models for point objects \cite[Chap. 5]{mahler2014}. 
We denote the state vector of object $i$ at time-step $t$ as $\mathbf{x}^t_i \in \mathbb R^{d_x}$, and the set of the states of all objects alive at time-step $t$ as $\mathbb X^t$. New objects appear according to a 
Poisson point process (PPP) parameterized with intensity function 
$\lambda_b(\mathbf x)$, while object death is modelled as independent and identically distributed (i.i.d.) Markovian processes, with survival probability $p_s(\mathbf x)$. The objects' motion models are also i.i.d.~Markovian processes, where the single-object transition density is denoted as $f(\mathbf x^{t+1} ~|~ \mathbf x^t)$.

The single-object measurement likelihood is denoted $\mathbf g(\mathbf z^t~|~\mathbf x^t)$, $\mathbf z^t\in\mathbb R^{d_z}$, where the probability of detection in state $\mathbf x$ is $p_d(\mathbf x)$ and each measurement is independent of all other objects and measurements conditioned on its corresponding target. Objects may generate at most one measurement per time-step, and measurements originate from at most one object. Clutter measurements are modeled as a PPP with constant intensity $\lambda_c$ over the field-of-view, and are independent of the existing objects and any other measurements. The set of all measurements generated at time-step $t$ (true measurements and clutter) is denoted $\mathbb{Z}^t$. 

\subsection{Problem formulation}
In this investigation, we focus on the problem of multi-object estimation using a sequence of measurements of arbitrary length, i.e., estimating $\mathbb X^T$ given access to measurements from $\tau$ time-steps in the past until the current time, i.e., $[\mathbb Z^{T-\tau}, \cdots, \mathbb Z^T]$. For applying a DL solution, we see this problem as a sequence-to-sequence mapping task, where a sequence of measurements $\mathbf z_{1:n}$ is to be mapped to a sequence $\mathbf y_{1:k}$.
The sequence $\mathbf z_{1:n}$ is formed by first appending each measurement vector in the moving window $[\mathbb Z^{T-\tau}, \cdots, \mathbb Z^T]$ with its time of measurement, and then joining all measurements into a single sequence, in arbitrary order. Hence, $n=\sum_{t=T-\tau}^T  |\mathbb{Z}^{t}|$. The sequence $\mathbf y_{1:k}$ specifies the predicted posterior density for $\mathbb X^T$ in the form of a Poisson multi-Bernoulli density\footnote
{
  A Poisson multi-Bernoulli density is the disjoint union of a PPP and a multi-Bernoulli (MB) density. In turn, an MB density is the disjoint union of Bernoulli components, each described by an existence probability and a state density function \cite{mahler2014}.
} with $k$ components. Each $\mathbf y_i\in\mathbb R^{d_y}$, $i\in\mathbb N_k$, contains the existence probability for that component and the parameters for describing its state density (e.g., mean and standard deviation).

\section{Background on Transformers}
\label{sec:background}
The DL method used in this paper is based on the Transformer architecture \cite{transformer_paper}, which in recent years has shown great potential in complex sequence-to-sequence function approximation \cite{transformers_for_translation, alphafold, transformers_in_tts, image_transformer}. 
This section provides a background on this type of neural network when processing an input sequence $\mathbf z_{1:n}$ with $\mathbf z_i\in \mathbb R^{d_z}$, $i\in\mathbb N_n$, to an output sequence $\mathbf y_{1:k}$ with $\mathbf y_i\in\mathbb R^{d_y}$, $i\in\mathbb N_k$. 

\subsection{Overall Architecture}
\begin{figure}
    \centering
    \tikzset{every picture/.style={line width=0.75pt}} 
        
    \begin{tikzpicture}[x=0.75pt,y=0.75pt,yscale=-1,xscale=1]
    
    \draw  [dash pattern={on 4.5pt off 4.5pt}] (72.13,162.96) .. controls (72.13,151.02) and (81.81,141.34) .. (93.75,141.34) -- (158.59,141.34) .. controls (170.52,141.34) and (180.2,151.02) .. (180.2,162.96) -- (180.2,312.65) .. controls (180.2,324.59) and (170.52,334.27) .. (158.59,334.27) -- (93.75,334.27) .. controls (81.81,334.27) and (72.13,324.59) .. (72.13,312.65) -- cycle ;
    \draw  [dash pattern={on 4.5pt off 4.5pt}] (233.36,77.76) .. controls (233.36,66.4) and (242.56,57.2) .. (253.91,57.2) -- (315.58,57.2) .. controls (326.93,57.2) and (336.13,66.4) .. (336.13,77.76) -- (336.13,314.24) .. controls (336.13,325.6) and (326.93,334.8) .. (315.58,334.8) -- (253.91,334.8) .. controls (242.56,334.8) and (233.36,325.6) .. (233.36,314.24) -- cycle ;
    \draw    (279.13,316.73) -- (279.13,294.3) ;
    \draw [shift={(279.13,291.3)}, rotate = 90] [fill={rgb, 255:red, 0; green, 0; blue, 0 }  ][line width=0.08]  [draw opacity=0] (7.14,-3.43) -- (0,0) -- (7.14,3.43) -- (4.74,0) -- cycle    ;
    \draw    (327.8,302.73) -- (279.47,302.73) ;
    \draw    (327.8,233.96) -- (327.8,302.73) ;
    \draw    (327.8,233.96) -- (316.02,233.96) ;
    \draw [shift={(313.02,233.96)}, rotate = 360] [fill={rgb, 255:red, 0; green, 0; blue, 0 }  ][line width=0.08]  [draw opacity=0] (7.14,-3.43) -- (0,0) -- (7.14,3.43) -- (4.74,0) -- cycle    ;
    \draw    (290.47,224.88) -- (290.47,205.54) ;
    \draw [shift={(290.47,202.54)}, rotate = 90] [fill={rgb, 255:red, 0; green, 0; blue, 0 }  ][line width=0.08]  [draw opacity=0] (7.14,-3.43) -- (0,0) -- (7.14,3.43) -- (4.74,0) -- cycle    ;
    \draw    (279.13,259.4) -- (279.13,246.54) ;
    \draw [shift={(279.13,243.54)}, rotate = 90] [fill={rgb, 255:red, 0; green, 0; blue, 0 }  ][line width=0.08]  [draw opacity=0] (7.14,-3.43) -- (0,0) -- (7.14,3.43) -- (4.74,0) -- cycle    ;
    \draw    (206.36,133.47) -- (206.36,214.04) ;
    \draw    (206.36,214.04) -- (269.02,214.04) ;
    \draw    (269.02,214.04) -- (269.02,205.54) ;
    \draw [shift={(269.02,202.54)}, rotate = 90] [fill={rgb, 255:red, 0; green, 0; blue, 0 }  ][line width=0.08]  [draw opacity=0] (7.14,-3.43) -- (0,0) -- (7.14,3.43) -- (4.74,0) -- cycle    ;
    \draw  [fill={rgb, 255:red, 255; green, 113; blue, 113 }  ,fill opacity=0.29 ] (244.47,227.81) .. controls (244.47,225.68) and (246.19,223.96) .. (248.33,223.96) -- (308.16,223.96) .. controls (310.29,223.96) and (312.02,225.68) .. (312.02,227.81) -- (312.02,239.39) .. controls (312.02,241.52) and (310.29,243.25) .. (308.16,243.25) -- (248.33,243.25) .. controls (246.19,243.25) and (244.47,241.52) .. (244.47,239.39) -- cycle ;
    \draw  [fill={rgb, 255:red, 186; green, 255; blue, 146 }  ,fill opacity=1 ] (244.69,266.71) .. controls (244.69,263.42) and (247.36,260.76) .. (250.64,260.76) -- (307.68,260.76) .. controls (310.97,260.76) and (313.63,263.42) .. (313.63,266.71) -- (313.63,284.58) .. controls (313.63,287.87) and (310.97,290.53) .. (307.68,290.53) -- (250.64,290.53) .. controls (247.36,290.53) and (244.69,287.87) .. (244.69,284.58) -- cycle ;
    \draw  [fill={rgb, 255:red, 186; green, 255; blue, 146 }  ,fill opacity=1 ] (240.36,179.38) .. controls (240.36,176.09) and (243.02,173.42) .. (246.31,173.42) -- (314.84,173.42) .. controls (318.13,173.42) and (320.8,176.09) .. (320.8,179.38) -- (320.8,197.24) .. controls (320.8,200.53) and (318.13,203.2) .. (314.84,203.2) -- (246.31,203.2) .. controls (243.02,203.2) and (240.36,200.53) .. (240.36,197.24) -- cycle ;
    \draw    (328.36,147.07) -- (328.36,162.96) -- (328.36,214.38) ;
    \draw    (328.36,147.07) -- (316.37,147.07) ;
    \draw [shift={(313.37,147.07)}, rotate = 360] [fill={rgb, 255:red, 0; green, 0; blue, 0 }  ][line width=0.08]  [draw opacity=0] (7.14,-3.43) -- (0,0) -- (7.14,3.43) -- (4.74,0) -- cycle    ;
    \draw    (278.47,173.4) -- (278.47,160.54) ;
    \draw [shift={(278.47,157.54)}, rotate = 90] [fill={rgb, 255:red, 0; green, 0; blue, 0 }  ][line width=0.08]  [draw opacity=0] (7.14,-3.43) -- (0,0) -- (7.14,3.43) -- (4.74,0) -- cycle    ;
    \draw  [fill={rgb, 255:red, 255; green, 113; blue, 113 }  ,fill opacity=0.29 ] (245.13,141.81) .. controls (245.13,139.68) and (246.86,137.96) .. (248.99,137.96) -- (308.83,137.96) .. controls (310.96,137.96) and (312.69,139.68) .. (312.69,141.81) -- (312.69,153.39) .. controls (312.69,155.52) and (310.96,157.25) .. (308.83,157.25) -- (248.99,157.25) .. controls (246.86,157.25) and (245.13,155.52) .. (245.13,153.39) -- cycle ;
    \draw    (290.47,214.38) -- (328.36,214.38) ;
    \draw    (276.8,138.29) -- (276.8,121.83) ;
    \draw [shift={(276.8,118.83)}, rotate = 90] [fill={rgb, 255:red, 0; green, 0; blue, 0 }  ][line width=0.08]  [draw opacity=0] (7.14,-3.43) -- (0,0) -- (7.14,3.43) -- (4.74,0) -- cycle    ;
    \draw    (277.8,98.97) -- (277.8,86.97) ;
    \draw [shift={(277.8,83.97)}, rotate = 90] [fill={rgb, 255:red, 0; green, 0; blue, 0 }  ][line width=0.08]  [draw opacity=0] (7.14,-3.43) -- (0,0) -- (7.14,3.43) -- (4.74,0) -- cycle    ;
    \draw  [fill={rgb, 255:red, 152; green, 249; blue, 255 }  ,fill opacity=1 ] (259.36,103.16) .. controls (259.36,101.02) and (261.09,99.29) .. (263.22,99.29) -- (292.43,99.29) .. controls (294.57,99.29) and (296.3,101.02) .. (296.3,103.16) -- (296.3,114.76) .. controls (296.3,116.89) and (294.57,118.62) .. (292.43,118.62) -- (263.22,118.62) .. controls (261.09,118.62) and (259.36,116.89) .. (259.36,114.76) -- cycle ;
    \draw  [fill={rgb, 255:red, 255; green, 113; blue, 113 }  ,fill opacity=0.29 ] (242.8,68.48) .. controls (242.8,66.35) and (244.53,64.62) .. (246.66,64.62) -- (306.5,64.62) .. controls (308.63,64.62) and (310.36,66.35) .. (310.36,68.48) -- (310.36,80.06) .. controls (310.36,82.19) and (308.63,83.92) .. (306.5,83.92) -- (246.66,83.92) .. controls (244.53,83.92) and (242.8,82.19) .. (242.8,80.06) -- cycle ;
    \draw    (326.02,72.4) -- (326.02,88.29) -- (326.02,131.96) ;
    \draw    (326.02,72.4) -- (313.87,72.4) ;
    \draw [shift={(310.87,72.4)}, rotate = 360] [fill={rgb, 255:red, 0; green, 0; blue, 0 }  ][line width=0.08]  [draw opacity=0] (7.14,-3.43) -- (0,0) -- (7.14,3.43) -- (4.74,0) -- cycle    ;
    \draw    (277.02,131.96) -- (326.02,131.96) ;
    \draw    (278.36,64.98) -- (278.36,38.76) ;
    \draw [shift={(278.36,35.76)}, rotate = 90] [fill={rgb, 255:red, 0; green, 0; blue, 0 }  ][line width=0.08]  [draw opacity=0] (7.14,-3.43) -- (0,0) -- (7.14,3.43) -- (4.74,0) -- cycle    ;
    \draw   (273.63,315.86) .. controls (273.63,312.82) and (276.1,310.36) .. (279.13,310.36) .. controls (282.17,310.36) and (284.63,312.82) .. (284.63,315.86) .. controls (284.63,318.9) and (282.17,321.36) .. (279.13,321.36) .. controls (276.1,321.36) and (273.63,318.9) .. (273.63,315.86) -- cycle ; \draw   (273.63,315.86) -- (284.63,315.86) ; \draw   (279.13,310.36) -- (279.13,321.36) ;
    \draw    (279.13,351.33) -- (279.13,323.73) ;
    \draw [shift={(279.13,320.73)}, rotate = 90] [fill={rgb, 255:red, 0; green, 0; blue, 0 }  ][line width=0.08]  [draw opacity=0] (7.14,-3.43) -- (0,0) -- (7.14,3.43) -- (4.74,0) -- cycle    ;
    \draw    (344.07,315.5) -- (287.63,315.5) ;
    \draw [shift={(284.63,315.5)}, rotate = 360] [fill={rgb, 255:red, 0; green, 0; blue, 0 }  ][line width=0.08]  [draw opacity=0] (7.14,-3.43) -- (0,0) -- (7.14,3.43) -- (4.74,0) -- cycle    ;
    \draw    (132.47,322.99) -- (132.47,293.46) ;
    \draw [shift={(132.47,290.46)}, rotate = 90] [fill={rgb, 255:red, 0; green, 0; blue, 0 }  ][line width=0.08]  [draw opacity=0] (7.14,-3.43) -- (0,0) -- (7.14,3.43) -- (4.74,0) -- cycle    ;
    \draw    (81.8,302.88) -- (132.47,302.88) ;
    \draw    (81.8,235.88) -- (81.8,302.88) ;
    \draw    (81.8,235.88) -- (97.37,235.88) ;
    \draw [shift={(100.37,235.88)}, rotate = 180] [fill={rgb, 255:red, 0; green, 0; blue, 0 }  ][line width=0.08]  [draw opacity=0] (7.14,-3.43) -- (0,0) -- (7.14,3.43) -- (4.74,0) -- cycle    ;
    \draw    (132.47,225.34) -- (132.47,208.01) ;
    \draw [shift={(132.47,205.01)}, rotate = 90] [fill={rgb, 255:red, 0; green, 0; blue, 0 }  ][line width=0.08]  [draw opacity=0] (7.14,-3.43) -- (0,0) -- (7.14,3.43) -- (4.74,0) -- cycle    ;
    \draw    (133.47,260.88) -- (133.47,248.03) ;
    \draw [shift={(133.47,245.03)}, rotate = 90] [fill={rgb, 255:red, 0; green, 0; blue, 0 }  ][line width=0.08]  [draw opacity=0] (7.14,-3.43) -- (0,0) -- (7.14,3.43) -- (4.74,0) -- cycle    ;
    \draw    (81.8,161.88) -- (96.44,161.88) ;
    \draw [shift={(99.44,161.88)}, rotate = 180] [fill={rgb, 255:red, 0; green, 0; blue, 0 }  ][line width=0.08]  [draw opacity=0] (7.14,-3.43) -- (0,0) -- (7.14,3.43) -- (4.74,0) -- cycle    ;
    \draw    (81.8,161.88) -- (81.8,217.18) ;
    \draw    (81.8,217.18) -- (132.47,217.18) ;
    \draw    (133.47,186.03) -- (133.47,174.03) ;
    \draw [shift={(133.47,171.03)}, rotate = 90] [fill={rgb, 255:red, 0; green, 0; blue, 0 }  ][line width=0.08]  [draw opacity=0] (7.14,-3.43) -- (0,0) -- (7.14,3.43) -- (4.74,0) -- cycle    ;
    \draw  [fill={rgb, 255:red, 186; green, 255; blue, 146 }  ,fill opacity=1 ] (99.02,266.63) .. controls (99.02,263.34) and (101.69,260.68) .. (104.98,260.68) -- (162.01,260.68) .. controls (165.3,260.68) and (167.97,263.34) .. (167.97,266.63) -- (167.97,284.5) .. controls (167.97,287.79) and (165.3,290.46) .. (162.01,290.46) -- (104.98,290.46) .. controls (101.69,290.46) and (99.02,287.79) .. (99.02,284.5) -- cycle ;
    \draw  [fill={rgb, 255:red, 255; green, 113; blue, 113 }  ,fill opacity=0.29 ] (100.47,229.2) .. controls (100.47,227.07) and (102.19,225.34) .. (104.33,225.34) -- (164.16,225.34) .. controls (166.29,225.34) and (168.02,227.07) .. (168.02,229.2) -- (168.02,240.78) .. controls (168.02,242.91) and (166.29,244.64) .. (164.16,244.64) -- (104.33,244.64) .. controls (102.19,244.64) and (100.47,242.91) .. (100.47,240.78) -- cycle ;
    \draw  [fill={rgb, 255:red, 152; green, 249; blue, 255 }  ,fill opacity=1 ] (115.02,190.21) .. controls (115.02,188.08) and (116.75,186.34) .. (118.89,186.34) -- (148.1,186.34) .. controls (150.24,186.34) and (151.97,188.08) .. (151.97,190.21) -- (151.97,201.81) .. controls (151.97,203.95) and (150.24,205.68) .. (148.1,205.68) -- (118.89,205.68) .. controls (116.75,205.68) and (115.02,203.95) .. (115.02,201.81) -- cycle ;
    \draw    (133.13,151.69) -- (133.13,144.34) -- (133.13,133.19) ;
    \draw    (133.13,133.19) -- (176.02,133.19) -- (206.36,133.19) ;
    \draw  [fill={rgb, 255:red, 255; green, 113; blue, 113 }  ,fill opacity=0.29 ] (99.47,155.54) .. controls (99.47,153.41) and (101.19,151.68) .. (103.33,151.68) -- (163.16,151.68) .. controls (165.29,151.68) and (167.02,153.41) .. (167.02,155.54) -- (167.02,167.12) .. controls (167.02,169.25) and (165.29,170.97) .. (163.16,170.97) -- (103.33,170.97) .. controls (101.19,170.97) and (99.47,169.25) .. (99.47,167.12) -- cycle ;
    \draw   (126.97,315.86) .. controls (126.97,312.82) and (129.43,310.36) .. (132.47,310.36) .. controls (135.5,310.36) and (137.97,312.82) .. (137.97,315.86) .. controls (137.97,318.9) and (135.5,321.36) .. (132.47,321.36) .. controls (129.43,321.36) and (126.97,318.9) .. (126.97,315.86) -- cycle ; \draw   (126.97,315.86) -- (137.97,315.86) ; \draw   (132.47,310.36) -- (132.47,321.36) ;
    \draw    (132.47,354.99) -- (132.47,324.99) ;
    \draw [shift={(132.47,321.99)}, rotate = 90] [fill={rgb, 255:red, 0; green, 0; blue, 0 }  ][line width=0.08]  [draw opacity=0] (7.14,-3.43) -- (0,0) -- (7.14,3.43) -- (4.74,0) -- cycle    ;
    \draw    (186.6,315.86) -- (140.97,315.86) ;
    \draw [shift={(137.97,315.86)}, rotate = 360] [fill={rgb, 255:red, 0; green, 0; blue, 0 }  ][line width=0.08]  [draw opacity=0] (7.14,-3.43) -- (0,0) -- (7.14,3.43) -- (4.74,0) -- cycle    ;
    
    \draw (335.13,59.7) node [anchor=north west][inner sep=0.75pt]    {$\times M$};
    \draw (249.97,228.5) node [anchor=north west][inner sep=0.75pt]  [font=\scriptsize,color={rgb, 255:red, 0; green, 0; blue, 0 }  ,opacity=1 ] [align=left] {Add \& Norm};
    \draw (252.63,265) node [anchor=north west][inner sep=0.75pt]  [font=\scriptsize,color={rgb, 255:red, 0; green, 0; blue, 0 }  ,opacity=1 ] [align=left] {\begin{minipage}[lt]{38.04pt}\setlength\topsep{0pt}
    \begin{center}
    Multi-head \\attention
    \end{center}
    
    \end{minipage}};
    \draw (246.31,177) node [anchor=north west][inner sep=0.75pt]  [font=\scriptsize,color={rgb, 255:red, 0; green, 0; blue, 0 }  ,opacity=1 ] [align=left] {\begin{minipage}[lt]{49.56pt}\setlength\topsep{0pt}
    \begin{center}
    Multi-head \\cross-attention
    \end{center}
    
    \end{minipage}};
    \draw (250.63,142.5) node [anchor=north west][inner sep=0.75pt]  [font=\scriptsize,color={rgb, 255:red, 0; green, 0; blue, 0 }  ,opacity=1 ] [align=left] {Add \& Norm};
    \draw (267.3,104.4) node [anchor=north west][inner sep=0.75pt]  [font=\scriptsize,color={rgb, 255:red, 0; green, 0; blue, 0 }  ,opacity=1 ] [align=left] {FFN};
    \draw (248.66,69.62) node [anchor=north west][inner sep=0.75pt]  [font=\scriptsize,color={rgb, 255:red, 0; green, 0; blue, 0 }  ,opacity=1 ] [align=left] {Add \& Norm};
    \draw (266,352.48) node [anchor=north west][inner sep=0.75pt]    {$\mathbf{o}_{1:k}$};
    \draw (266.47,20) node [anchor=north west][inner sep=0.75pt]    {$\mathbf{y}_{1:k}$};
    \draw (189.4,308) node [anchor=north west][inner sep=0.75pt]    {$\mathbf{q}_{1:n}^{e}$};
    \draw (347,306) node [anchor=north west][inner sep=0.75pt]    {$\mathbf{q}_{1:k}^{d}$};
    \draw (106.98,265) node [anchor=north west][inner sep=0.75pt]  [font=\scriptsize,color={rgb, 255:red, 0; green, 0; blue, 0 }  ,opacity=1 ] [align=left] {\begin{minipage}[lt]{38.04pt}\setlength\topsep{0pt}
    \begin{center}
    Multi-head \\attention
    \end{center}
    
    \end{minipage}};
    \draw (105.97,229.88) node [anchor=north west][inner sep=0.75pt]  [font=\scriptsize,color={rgb, 255:red, 0; green, 0; blue, 0 }  ,opacity=1 ] [align=left] {Add \& Norm};
    \draw (122.89,191.34) node [anchor=north west][inner sep=0.75pt]  [font=\scriptsize,color={rgb, 255:red, 0; green, 0; blue, 0 }  ,opacity=1 ] [align=left] {FFN};
    \draw (46,159.52) node [anchor=north west][inner sep=0.75pt]    {$N\times $};
    \draw (120.93,354.6) node [anchor=north west][inner sep=0.75pt]    {$\mathbf{z}_{1:n}$};
    \draw (156.8,120) node [anchor=north west][inner sep=0.75pt]    {$\mathbf{e}_{1:n}$};
    \draw (103.97,156.22) node [anchor=north west][inner sep=0.75pt]  [font=\scriptsize,color={rgb, 255:red, 0; green, 0; blue, 0 }  ,opacity=1 ] [align=left] {Add \& Norm};

    \end{tikzpicture}

    \caption{Simplified diagram illustrating the Transformer architecture. Encoder on the left, containing $N$ encoder blocks, processes the input sequence $\mathbf z_{1:n}$ into embeddings $\mathbf e_{1:n}$. Decoder on the right, containing $M$ decoder blocks, uses the embeddings $\mathbf e_{1:n}$ produced by the encoder together with the object queries $\mathbf o_{1:k}$ to predict the output sequence $\mathbf y_{1:k}$. FFN stands for fully-connected feedforward neural network.}
    \label{fig:encoder-decoder_diagram}
    \vspace{-2mm}
\end{figure}
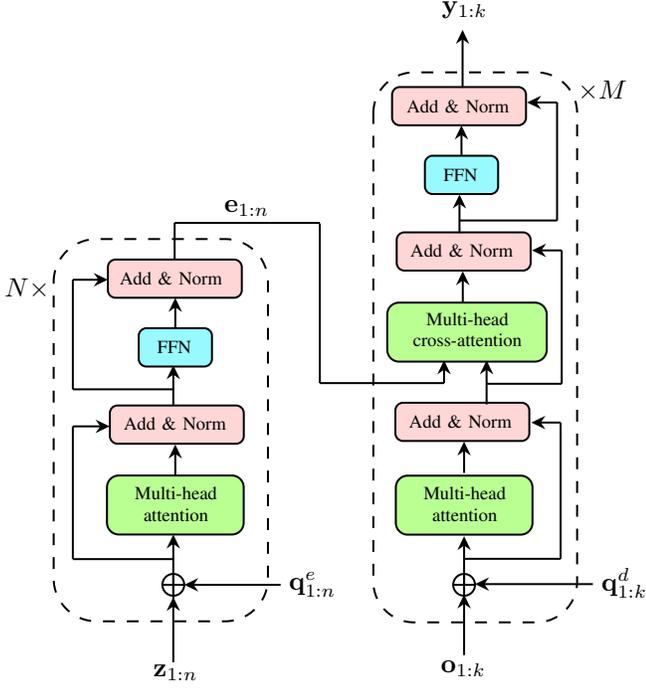

The Transformer architecture is comprised of two main components: an encoder and a decoder, as depicted in Fig.\,\ref{fig:encoder-decoder_diagram}. The encoder is in charge of processing the input sequence $\mathbf z_{1:n}$ so that each element is transformed into a new representation that encodes both its value and its relationship to other elements of the sequence. This is accomplished primarily by the use of a special type of layer called \textit{self-attention}, and the new sequence is referred to as the embeddings of the input sequence, $\mathbf e_{1:n}$, with $\mathbf e_i\in\mathbb R^{d_e}$, $i\in\mathbb N_n$. The decoder then processes these embeddings (using slightly modified self-attention layers) into an output sequence $\mathbf y_{1:k}$, either autoregressively \cite{transformer_paper} or using learned input queries $\mathbf o_{1:k}$ \cite{DETR}. These components together make for a powerful learnable mapping between an input sequence $\mathbf z_{1:n}$ and an output sequence $\mathbf y_{1:k}$, typically trained using stochastic gradient descent on a loss function $\mathcal L(\mathbf y_{1:k}, \mathbf x_{1:k})$ that compares the network predictions with a ground-truth sequence $\mathbf x_{1:k}$.

\subsection{Multi-head Self-attention Layer}
\label{subsec:multi-head_self_attention_layer}
The main building block for the Transformer architecture is the \textit{self-attention layer}, used multiple times inside both the encoder and decoder modules. Since it is used in different places in the Transformer architecture, we describe it here generally, as a learnable mapping between an input sequence $\mathbf a_{1:n}$ and an output sequence $\mathbf b_{1:n}$, where $\mathbf a_i, \mathbf b_i\in\mathbb R^d, i\in\mathbb N_n$. The self-attention layer first computes three different linear transformations of the input:
\begin{equation}
    \label{eq:self-attention_qkv}
    \mathbf{Q} = \mathbf{W}_Q \mathbf{A},~ \mathbf{K} = \mathbf{W}_K\mathbf{A},~ \mathbf{V}=\mathbf{W}_V\mathbf{A},
\end{equation}
where $\mathbf{A} =\begin{bmatrix}\mathbf a_1, \cdots, \mathbf a_n\end{bmatrix} \in \mathbb R^{d\times n}$, and the matrices $\mathbf Q, \mathbf K, \mathbf V$ are referred to as queries, keys, and values, respectively. The matrices $\mathbf{W}_Q, \mathbf{W}_K, \mathbf{W}_V \in \mathbb R^{d\times d}$ are the learnable parameters of the self-attention layer. The output is then computed as
\begin{equation}
    \mathbf{B}=\mathbf{V}\cdot  \text{Softmax-c}\left(\frac{\mathbf{K}^\top\mathbf{Q}}{\sqrt{d}}\right)~,
\end{equation}
where $\mathbf{B} = \begin{bmatrix}\mathbf b_1, \cdots, \mathbf b_n\end{bmatrix}\in \mathbb R^{d\times n}$ and Softmax-c is the column-wise application of the Softmax function, defined as
\begin{align*}
    &[\text{Softmax-c}(\mathbf Z)]_{i,j} = \frac{e^{z_{i,j}}}{\sum_{k=1}^{d} e^{z_{k,j}}};\quad i, j\in\mathbb N_n
\end{align*} 
for $\mathbf{Z}\in \mathbb R^{n\times n}$, where $z_{i,j}$ is the element of $\mathbf Z$ on row $i$, column $j$. Because of this structure, each output $\mathbf b_i$ from a self-attention layer directly depends on all inner-products of the type $\mathbf a_i^\top \mathbf W\mathbf a_j$, for $j\in\mathbb N_n$, with learnable $\mathbf W$, between the elements of the input sequence. This allows for the potential to learn an improved representation of each $\mathbf a_i$ that takes into account its relationship to all the other elements of the sequence. Compound applications of these layers can then result in complex representations of the input sequence that take into account more complicated relationships between all the elements. This property is potentially very helpful in MOT, allowing the model to learn and exploit complicated, long-range patterns in the sequence of measurements when tracking objects.

In practice, Transformer-based architectures often use several self-attention layers in parallel and then combine the results, where this entire computation is referred to as a multi-head self-attention layer (shown in green in Fig.\,\ref{fig:encoder-decoder_diagram}). For this, $\mathbf A$ is fed to $n_h$ different self-attention layers (with separate learnable parameters) in parallel, generating $n_h$ different outputs $\mathbf{B}_1, \cdots, \mathbf{B}_{n_h}$, all $\in\mathbb R^{d\times n}$. The final output $\mathbf B$ is then computed by vertically stacking the results and applying a linear transformation to reduce the dimensionality back to $\mathbb R^{d\times n}$:
\begin{align}
    \mathbf{B} &= \mathbf{W}^0\begin{bmatrix}
        \mathbf{B}_1 \\
        \vdots       \\
        \mathbf{B}_{n_h}\end{bmatrix}
\end{align}
where $\mathbf{W}^0\in\mathbb R^{d\times dn_n}$ is a learnable parameter of the multi-head self-attention layer. Finally, $\mathbf{B}$ is converted back to a sequence $\mathbf{b}_{1:n}=\mathrm{MultiHeadAttention}(\mathbf{a}_{1:n})$.

\subsection{Transformer Encoder}
The Transformer encoder is the module in charge of transforming the input sequence $\mathbf z_{1:n}$ into the embeddings $\mathbf e_{1:n}$, where, after training, element $\mathbf e_i$ can potentially encode both the original value $\mathbf z_i$ and any important relationships it has to the other elements in the sequence. In MOT, for example, $\mathbf e_i$ can contain relevant information about other measurements that originate from the same object as $\mathbf z_i$.

A Transformer encoder is built from $N$ encoder blocks in series, as shown in the left of Fig.\,\ref{fig:encoder-decoder_diagram}. The output for encoder block $l\in\mathbb N_N$ is computed as 
\begin{align}
    \label{eq:encoder_eq1}
    &\tilde{\mathbf z}_{1:n}^{(l-1)} = \mathbf z_{1:n}^{(l-1)} + \mathbf q_{1:n}^e
    \\
    \label{eq:encoder_eq2}
    &\mathbf t^{(l)}_{1:n} = \mathrm{MultiHeadAttention}(\tilde{\mathbf z}_{1:n}^{(l-1)})
    \\
    \label{eq:encoder_eq3}
    &\tilde{\mathbf{t}}^{(l)}_{1:n} = \mathrm{LayerNorm}(\tilde{\mathbf z}_{1:n}^{(l-1)} + \mathbf t^{(l)}_{1:n})
    \\
    \label{eq:encoder_eq4}
    &\mathbf z_{1:n}^{(l)} = \mathrm{LayerNorm}(
    \tilde{\mathbf{t}}^{(l)}_{1:n} + \text{FFN}(\tilde{\mathbf{t}}^{(l)}_{1:n}))~,
\end{align}
where MultiHeadAttention is a multi-head self-attention layer, as described in Section \ref{subsec:multi-head_self_attention_layer}, FFN is a fully-connected feedforward neural network applied to each element of the input sequence separately, LayerNorm is a Layer Normalization layer (as introduced in \cite{layer_normalization}), and $\mathbf z_{1:n}^{(l)}$ is the input sequence after being processed by $l$ encoder blocks. 
Hence, $\mathbf z_{1:n}^{(0)}$ is the original input sequence $\mathbf z_{1:n}$, and $\mathbf z_{1:n}^{(N)}$ is the output of the encoder module, also denoted $\mathbf e_{1:n}$. Note that both multi-head self-attention and layer normalization preserve the size of the input, so $\mathbf z_{1:n}^{(l)}\in\mathbb R^{d_z}$, for $l\in\mathbb N_N$ 
Importantly, $\mathbf q_{1:n}^e$ in \eqref{eq:encoder_eq1}, referred to as the positional encoding for the input sequence, is added to the input of every encoder block (as done in \cite{DETR}), computed
as $\mathbf q_{i}^e=f_p^e(i)$, where $f: \mathbb{Z} \to \mathbb{R}^{d_z}$, and $f_p^e$ can either be fixed (usually with sinusoidal components \cite{transformer_paper}) or learnable \cite{DETR}. Without it, the encoder module becomes permutation-equivariant\footnote{A function $f$ is equivariant to a transformation $g$ iff $f(g(x))=g(f(x)).$}, which is undesirable in many contexts. For instance, when processing images with Transformers the order of the elements in the input sequence is related to their location in the image, and therefore very important for correctly solving non-trivial tasks.

\subsection{Transformer Decoder}
Once the embeddings $\mathbf e_{1:n}$ are computed by the encoder, the decoder module is in charge of using them to predict the output sequence $\mathbf y_{1:k}$. Different structures for the Transformer decoder have been proposed for different contexts \cite{transformer_paper, evolved_transformer, decoder_for_speech_recon_and_translation}, and the one used for this paper is based on the DEtection TRansformer (DETR) decoder \cite{DETR} using object queries $\mathbf o_{1:k}$ (illustrated on the right part of Fig.\,\ref{fig:encoder-decoder_diagram}), due to its speed and capacity to generate outputs in parallel, instead of autoregressively. This type of decoder, just like the encoder module, is comprised of $M$ decoder blocks, where the output for decoder block $l\in\mathbb N_M$ is computed as
\begin{align}
    &\tilde{\mathbf o}_{1:k}^{(l-1)} = \mathbf o_{1:k}^{(l-1)} + \mathbf q_{1:k}^d
    \\
    \label{eq:decoder_eq1}
    &\mathbf r^{(l)}_{1:k} = \mathrm{MultiHeadAttention}(\tilde{\mathbf o}_{1:k}^{(l-1)})
    \\
    &\tilde{\mathbf{r}}^{(l)}_{1:k} = \mathrm{LayerNorm}(\tilde{\mathbf o}_{1:k}^{(l-1)} + \mathbf r^{(l)}_{1:k})
    \\
    &\tilde{\mathbf e}^{(l)}_{1:k} = \mathrm{MultiHeadCrossAttention}(\tilde{\mathbf r}_{1:k}^{(l)}, \mathbf e_{1:n})
    \\
    &\bar{\mathbf e}_{1:k}^{(l)} = \mathrm{LayerNorm}(
    \tilde{\mathbf{r}}^{(l)}_{1:k} + \tilde{\mathbf e}^{(l)}_{1:k}))
    \\
    &\mathbf o_{1:k}^{(l)} = \mathrm{LayerNorm}(\bar{\mathbf e}_{1:k}^{(l)} + \text{FFN}(\bar{\mathbf e}_{1:k}^{(l)})~,
\end{align}
where $\mathrm{MultiHeadCrossAttention}$ is a regular multi-head self-attention layer as described in Section \ref{subsec:multi-head_self_attention_layer}, with the difference that the matrices $\mathbf K$, $\mathbf Q$, and $\mathbf V$ in \eqref{eq:self-attention_qkv} are respectively computed as $\mathbf W_K\mathbf e_{1:n}$, $\mathbf W_Q \tilde{\mathbf r}_{1:k}^{(l)}$, and $\mathbf W_V\mathbf e_{1:n}$ (all of the subsequent self-attention computations are the same).
The input to the first encoder block are the object queries $\mathbf o_{1:k}$, a sequence of learnable vectors trained jointly with the other model parameters. Once trained, each $\mathbf o_i\in\mathbb R^{d_o}$, $i\in\mathbb N_k$, will potentially learn to attend to the parts of the embeddings $\mathbf e_{1:n}$ that are helpful for predicting $\mathbf y_i$. Similar to the encoder module, $\mathbf o_{1:k}^{(l)}$ represents the object queries after being processed by $l$ decoder blocks (which also preserve the size of the input), where $\mathbf o_{1:k}^{(M)}$ denotes the output of the decoder module $\mathbf y_{1:k}$. Finally, to prevent the decoder module from being permutation-equivariant, a positional encoding $\mathbf q_{1:k}^d$ is added to the inputs of each layer, where $\mathbf q_i^d=f_p^d(i)$.

\section{MultiTarget Tracking Transformer v2}
\label{sec:mt3v2}

The state of the art in DL-based model-free tracking is often tailored for high-dimensional or dense inputs such as images or LIDAR clouds \cite{model-free_mot2, model-free_mot3, model-free_mot4, model-free_mot5, model-free_mot1}, and therefore relies on network architectures comprised of stacks of layers with inductive biases tailored for this type of data (e.g.,  convolutional neural network layers, voxel feature encoding layers \cite{vfe_layer}, etc). Although possible to apply such methods to the model-based, low-dimensional setting, we expect their performance to be suboptimal (nor efficient in terms of training time) unless considerable work is devoted to tailoring their architectures for the problem at hand. Therefore, instead of attempting to adapt and benchmark model-free DL trackers in the model-based setting, we propose a novel, high-performing architecture tailored for the model-based MOT context, and use it as proof of concept for the potential of DL in model-based MOT.

As mentioned in Section \ref{sec:problem_formulation}, we see the model-based MOT problem as the task of mapping a sequence of measurements $\mathbf z_{1:n}$ to a sequence of predictions $\mathbf y_{1:k}$, corresponding to the parameters of a multi-Bernoulli density (state distribution and existence probability for each component) describing the objects present at time-step $T$. Using the available transition and measurement models of the environment, we generate unlimited training data to train MT3v2 to learn this mapping. By approaching the problem as a sequence-to-sequence learning problem using a Transformer-based model, we are able to train a tracker that uses a constant number of parameters regardless of the number of time-steps being processed. In this way, we sidestep the need for using heuristics to maintaining computational tractability that often impact performance.

\subsection{Overview of MT3v2 architecture}

\begin{figure}[t]
    \centering

\tikzset{every picture/.style={line width=0.75pt}} 

\begin{tikzpicture}[x=0.75pt,y=0.75pt,yscale=-1,xscale=1]

\draw  [fill={rgb, 255:red, 206; green, 219; blue, 255 }  ,fill opacity=1 ][dash pattern={on 4.5pt off 4.5pt}] (32.33,182.75) .. controls (32.33,166.43) and (45.56,153.2) .. (61.88,153.2) -- (165.22,153.2) .. controls (181.54,153.2) and (194.77,166.43) .. (194.77,182.75) -- (194.77,271.39) .. controls (194.77,287.71) and (181.54,300.94) .. (165.22,300.94) -- (61.88,300.94) .. controls (45.56,300.94) and (32.33,287.71) .. (32.33,271.39) -- cycle ;
\draw  [fill={rgb, 255:red, 174; green, 167; blue, 255 }  ,fill opacity=1 ] (88.03,254.33) .. controls (88.03,250.87) and (90.84,248.07) .. (94.3,248.07) -- (133.8,248.07) .. controls (137.26,248.07) and (140.07,250.87) .. (140.07,254.33) -- (140.07,273.13) .. controls (140.07,276.59) and (137.26,279.4) .. (133.8,279.4) -- (94.3,279.4) .. controls (90.84,279.4) and (88.03,276.59) .. (88.03,273.13) -- cycle ;

\draw  [fill={rgb, 255:red, 184; green, 255; blue, 165 }  ,fill opacity=1 ] (48.4,174.76) .. controls (48.4,170.51) and (51.84,167.07) .. (56.09,167.07) -- (98.84,167.07) .. controls (103.09,167.07) and (106.53,170.51) .. (106.53,174.76) -- (106.53,197.84) .. controls (106.53,202.09) and (103.09,205.53) .. (98.84,205.53) -- (56.09,205.53) .. controls (51.84,205.53) and (48.4,202.09) .. (48.4,197.84) -- cycle ;

\draw  [fill={rgb, 255:red, 174; green, 167; blue, 255 }  ,fill opacity=1 ] (130.97,177.6) .. controls (130.97,174.14) and (133.77,171.33) .. (137.23,171.33) -- (176.73,171.33) .. controls (180.19,171.33) and (183,174.14) .. (183,177.6) -- (183,196.4) .. controls (183,199.86) and (180.19,202.67) .. (176.73,202.67) -- (137.23,202.67) .. controls (133.77,202.67) and (130.97,199.86) .. (130.97,196.4) -- cycle ;

\draw    (157.81,224.14) -- (86,224.14) -- (86,208.54) ;
\draw [shift={(86,205.54)}, rotate = 90] [fill={rgb, 255:red, 0; green, 0; blue, 0 }  ][line width=0.08]  [draw opacity=0] (8.93,-4.29) -- (0,0) -- (8.93,4.29) -- cycle    ;
\draw    (158.01,264.73) -- (158.01,205.05) ;
\draw [shift={(158.01,202.05)}, rotate = 90] [fill={rgb, 255:red, 0; green, 0; blue, 0 }  ][line width=0.08]  [draw opacity=0] (8.93,-4.29) -- (0,0) -- (8.93,4.29) -- cycle    ;
\draw    (107.1,187.67) -- (127.77,187.67) ;
\draw [shift={(130.77,187.67)}, rotate = 180] [fill={rgb, 255:red, 0; green, 0; blue, 0 }  ][line width=0.08]  [draw opacity=0] (8.93,-4.29) -- (0,0) -- (8.93,4.29) -- cycle    ;
\draw  [fill={rgb, 255:red, 255; green, 217; blue, 144 }  ,fill opacity=1 ][dash pattern={on 4.5pt off 4.5pt}] (233.36,297.29) .. controls (220.89,297.26) and (210.8,287.12) .. (210.83,274.64) -- (211.12,162.65) .. controls (211.16,150.17) and (221.3,140.08) .. (233.77,140.12) -- (301.55,140.29) .. controls (314.03,140.33) and (324.12,150.47) .. (324.08,162.94) -- (323.79,274.93) .. controls (323.76,287.41) and (313.62,297.5) .. (301.14,297.47) -- cycle ;
\draw  [fill={rgb, 255:red, 255; green, 176; blue, 0 }  ,fill opacity=1 ] (232.51,257.14) .. controls (232.56,254.27) and (234.93,252) .. (237.79,252.05) -- (261.68,252.52) .. controls (264.55,252.58) and (266.82,254.95) .. (266.77,257.81) -- (266.46,273.37) .. controls (266.4,276.23) and (264.04,278.51) .. (261.17,278.45) -- (237.28,277.98) .. controls (234.42,277.92) and (232.14,275.56) .. (232.2,272.69) -- cycle ;

\draw    (248.44,314.12) -- (248.53,281.19) ;
\draw [shift={(248.53,278.19)}, rotate = 90.15] [fill={rgb, 255:red, 0; green, 0; blue, 0 }  ][line width=0.08]  [draw opacity=0] (8.93,-4.29) -- (0,0) -- (8.93,4.29) -- cycle    ;
\draw  [fill={rgb, 255:red, 255; green, 176; blue, 0 }  ,fill opacity=1 ] (252.07,153.34) .. controls (252.06,150.47) and (254.37,148.14) .. (257.24,148.13) -- (281.13,148.03) .. controls (283.99,148.02) and (286.33,150.33) .. (286.34,153.2) -- (286.4,168.76) .. controls (286.41,171.62) and (284.1,173.95) .. (281.23,173.97) -- (257.34,174.06) .. controls (254.48,174.07) and (252.14,171.76) .. (252.13,168.9) -- cycle ;

\draw  [fill={rgb, 255:red, 255; green, 176; blue, 0 }  ,fill opacity=1 ] (251.09,209.35) .. controls (251.06,206.49) and (253.36,204.15) .. (256.23,204.12) -- (280.12,203.9) .. controls (282.98,203.88) and (285.33,206.18) .. (285.35,209.04) -- (285.49,224.6) .. controls (285.52,227.47) and (283.22,229.81) .. (280.35,229.84) -- (256.46,230.05) .. controls (253.6,230.08) and (251.25,227.78) .. (251.23,224.91) -- cycle ;

\draw    (268.9,174.38) -- (268.9,199.38) ;
\draw [shift={(268.9,202.38)}, rotate = 270] [fill={rgb, 255:red, 0; green, 0; blue, 0 }  ][line width=0.08]  [draw opacity=0] (8.93,-4.29) -- (0,0) -- (8.93,4.29) -- cycle    ;
\draw    (304.26,215.93) -- (288.66,215.89) ;
\draw [shift={(285.66,215.88)}, rotate = 0.15] [fill={rgb, 255:red, 0; green, 0; blue, 0 }  ][line width=0.08]  [draw opacity=0] (8.93,-4.29) -- (0,0) -- (8.93,4.29) -- cycle    ;
\draw    (304.26,215.93) -- (304.4,161.93) ;
\draw    (304.4,161.93) -- (288.8,161.89) ;
\draw [shift={(285.8,161.88)}, rotate = 0.15] [fill={rgb, 255:red, 0; green, 0; blue, 0 }  ][line width=0.08]  [draw opacity=0] (8.93,-4.29) -- (0,0) -- (8.93,4.29) -- cycle    ;
\draw    (304.33,188.93) -- (330.46,189) ;
\draw    (233.93,161.36) -- (249.53,161.36) ;
\draw [shift={(252.53,161.36)}, rotate = 180] [fill={rgb, 255:red, 0; green, 0; blue, 0 }  ][line width=0.08]  [draw opacity=0] (8.93,-4.29) -- (0,0) -- (8.93,4.29) -- cycle    ;
\draw    (233.93,161.36) -- (233.93,215.36) ;
\draw    (233.93,215.36) -- (249.53,215.36) ;
\draw [shift={(252.53,215.36)}, rotate = 180] [fill={rgb, 255:red, 0; green, 0; blue, 0 }  ][line width=0.08]  [draw opacity=0] (8.93,-4.29) -- (0,0) -- (8.93,4.29) -- cycle    ;
\draw    (233.93,187.36) -- (183.57,187.36) ;
\draw  [fill={rgb, 255:red, 255; green, 255; blue, 255 }  ,fill opacity=1 ] (299.53,271.83) .. controls (295.83,271.82) and (292.83,268.81) .. (292.84,265.11) .. controls (292.85,261.41) and (295.86,258.42) .. (299.56,258.43) .. controls (303.26,258.44) and (306.25,261.45) .. (306.24,265.15) .. controls (306.23,268.85) and (303.23,271.84) .. (299.53,271.83) -- cycle ; \draw   (299.53,271.83) -- (299.56,258.43) ; \draw   (292.84,265.11) -- (306.24,265.15) ;
\draw    (266.44,265.05) -- (289.84,265.05) ;
\draw [shift={(292.84,265.05)}, rotate = 180] [fill={rgb, 255:red, 0; green, 0; blue, 0 }  ][line width=0.08]  [draw opacity=0] (8.93,-4.29) -- (0,0) -- (8.93,4.29) -- cycle    ;
\draw    (299.53,271.83) -- (299.53,313.83) ;
\draw [shift={(299.53,316.83)}, rotate = 270] [fill={rgb, 255:red, 0; green, 0; blue, 0 }  ][line width=0.08]  [draw opacity=0] (8.93,-4.29) -- (0,0) -- (8.93,4.29) -- cycle    ;
\draw    (269.29,230.71) -- (269.29,241.38) -- (299.43,241.38) -- (299.54,255.43) ;
\draw [shift={(299.56,258.43)}, rotate = 269.55] [fill={rgb, 255:red, 0; green, 0; blue, 0 }  ][line width=0.08]  [draw opacity=0] (8.93,-4.29) -- (0,0) -- (8.93,4.29) -- cycle    ;
\draw    (139.8,264.73) -- (229.76,264.73) ;
\draw [shift={(232.76,264.73)}, rotate = 180] [fill={rgb, 255:red, 0; green, 0; blue, 0 }  ][line width=0.08]  [draw opacity=0] (8.93,-4.29) -- (0,0) -- (8.93,4.29) -- cycle    ;
\draw    (66.2,312.14) -- (66.98,208.54) ;
\draw [shift={(67,205.54)}, rotate = 90.43] [fill={rgb, 255:red, 0; green, 0; blue, 0 }  ][line width=0.08]  [draw opacity=0] (8.93,-4.29) -- (0,0) -- (8.93,4.29) -- cycle    ;
\draw    (66.6,263.67) -- (85.77,263.67) ;
\draw [shift={(88.77,263.67)}, rotate = 180] [fill={rgb, 255:red, 0; green, 0; blue, 0 }  ][line width=0.08]  [draw opacity=0] (8.93,-4.29) -- (0,0) -- (8.93,4.29) -- cycle    ;

\draw (105.13,137) node [anchor=north west][inner sep=0.75pt]  [font=\footnotesize] [align=left] {\textit{MT3v2}};
\draw (94,257.2) node [anchor=north west][inner sep=0.75pt]  [font=\footnotesize] [align=left] {Encoder};
\draw (50.73,175) node [anchor=north west][inner sep=0.75pt]  [font=\scriptsize] [align=left] {\begin{minipage}[lt]{39.22pt}\setlength\topsep{0pt}
\begin{center}
Selection\\mechanism
\end{center}

\end{minipage}};
\draw (137,180.46) node [anchor=north west][inner sep=0.75pt]  [font=\footnotesize] [align=left] {Decoder};
\draw (54.2,314.9) node [anchor=north west][inner sep=0.75pt]  [font=\footnotesize]  {$\mathbf{z}_{1:n}$};
\draw (166.4,251) node [anchor=north west][inner sep=0.75pt]  [font=\footnotesize]  {$\mathbf{e}_{1:n}$};
\draw (239,314) node [anchor=north west][inner sep=0.75pt]  [font=\footnotesize,rotate=-0.26]  {$b_{1:n}$};
\draw (254.72,124.8) node [anchor=north west][inner sep=0.75pt]  [font=\footnotesize,rotate=-359.57] [align=left] {\begin{minipage}[lt]{19.96pt}\setlength\topsep{0pt}
\begin{center}
\textit{Loss}
\end{center}

\end{minipage}};
\draw (275,317.19) node [anchor=north west][inner sep=0.75pt]  [font=\footnotesize,rotate=-359.84]  {$\text{Total Loss}$};
\draw (333.81,184) node [anchor=north west][inner sep=0.75pt]  [font=\footnotesize,rotate=-0.48]  {$\mathbf{x}_{1:m}$};
\draw (264.23,211.05) node [anchor=north west][inner sep=0.75pt]  [font=\scriptsize,rotate=-359.48]  {$\mathcal{L}$};
\draw (253.73,154.28) node [anchor=north west][inner sep=0.75pt]  [font=\scriptsize,rotate=-359.77]  {$\mathcal{L}_{\text{match}}$};
\draw (242.12,259.27) node [anchor=north west][inner sep=0.75pt]  [font=\scriptsize,rotate=-1.13]  {$\mathcal{L}_{\text{c}}$};

\end{tikzpicture}

    \caption{Overview of the MT3v2 architecture. Input sequence of measurements $\mathbf z_{1:n}$ is processed by the encoder, generating the embeddings $\mathbf e_{1:n}$ and by the selection mechanism, generating the initial estimates $\tilde{\mathbf z}_{1:k}$, object queries $\mathbf o_{1:k}$, and positional encodings $\mathbf q_{1:k}^d$ for the decoder. The embeddings from the encoder, along with the output of the selection mechanism, are used by the decoder to output $\mathbf y_{1:k}$, describing a multi-Bernoulli density with $k$ components.}
    \label{fig:mt3v2_orverview}
    \vspace{-2mm}
\end{figure}
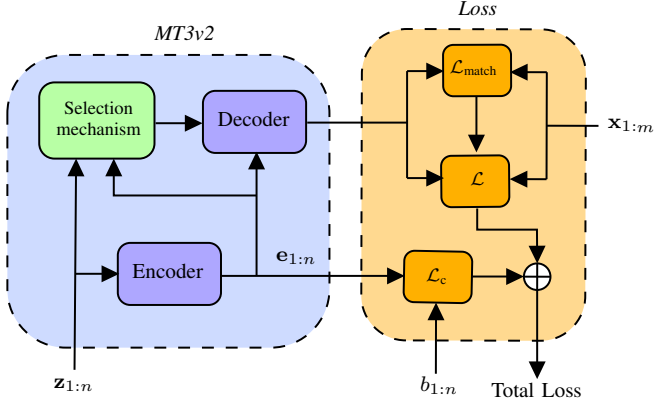

The MT3v2 architecture is comprised of a Transformer encoder, a modified Transformer decoder, and a selection mechanism, as shown in the left of Fig.\,\ref{fig:mt3v2_orverview}. The idea behind this specific structure is that the encoder can process the measurement sequence into a new representation $\mathbf e_{1:n}$ that summarizes relevant information to the MOT task, such as which are the clutter measurements, which measurements come from the same objects, etc. Then, instead of using a decoder with object queries which are independent of the input measurements (and therefore forced to be general), the selection mechanism uses the generated embeddings and measurements to create object queries $\mathbf o_{1:k}$ (and corresponding positional embeddings) which are specifically suited for the current sequence $\mathbf z_{1:n}$. 
Furthermore, in order to relieve the decoder from the burden of having to generate predictions from scratch, the selection mechanism also generates potential starting points $\tilde{\mathbf z}_{1:k}$, which the decoder then iteratively refines (predicts additive adjustments) at each decoder block. The final output sequence from the decoder, denoted $\mathbf y_{1:k}$, represents the parameters of a $k$-component MB density. Each $\mathbf y_i, i\in\mathbb N_k$ is of the form $(\boldsymbol\mu_i, \boldsymbol\Sigma_i, p_i)$, containing respectively the mean and covariance for a Gaussian distribution, and the existence probability for that Bernoulli component.

Both the output sequence $\mathbf y_{1:k}$ from the decoder (along with outputs from the intermediate decoder blocks, see section \ref{subsec:loss}) and the embeddings $\mathbf e_{1:n}$ are used for training MT3v2. The output sequence is used to approximate the negative log-likelihood of the predicted multi-Bernoulli densities, while the embeddings are used for computing an auxiliary contrastive loss that accelerates learning. Training is then performed by optimizing the sum of these two different losses.

The rest of this section explains the selection mechanism and the iterative refinement process in the decoder in more detail, followed by a description of the negative log-likelihood loss and the contrastive auxiliary loss used, and finalizes with information about the most important preprocessing steps applied to the training data.

\subsection{Selection Mechanism}
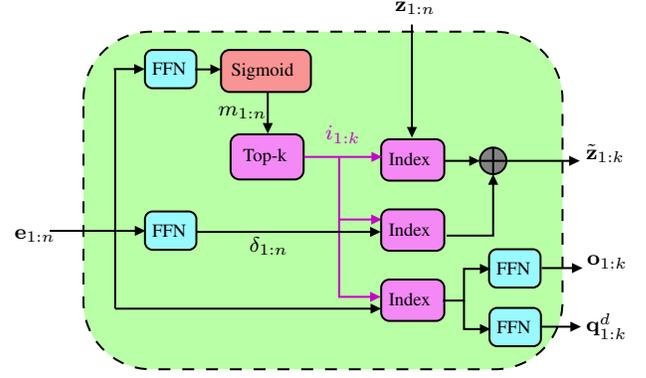
\begin{figure}[t]
    \centering

\tikzset{every picture/.style={line width=0.75pt}} 

\begin{tikzpicture}[x=0.75pt,y=0.75pt,yscale=-1,xscale=1]

\draw  [fill={rgb, 255:red, 184; green, 255; blue, 165 }  ,fill opacity=1 ][dash pattern={on 4.5pt off 4.5pt}] (121.25,106.74) .. controls (121.25,87.92) and (136.51,72.66) .. (155.33,72.66) -- (328.78,72.66) .. controls (347.6,72.66) and (362.86,87.92) .. (362.86,106.74) -- (362.86,208.98) .. controls (362.86,227.8) and (347.6,243.06) .. (328.78,243.06) -- (155.33,243.06) .. controls (136.51,243.06) and (121.25,227.8) .. (121.25,208.98) -- cycle ;
\draw    (137.91,173.2) -- (149.53,173.2) ;
\draw [shift={(152.53,173.2)}, rotate = 180] [fill={rgb, 255:red, 0; green, 0; blue, 0 }  ][line width=0.08]  [draw opacity=0] (5.36,-2.57) -- (0,0) -- (5.36,2.57) -- cycle    ;
\draw    (214.2,103) -- (214.2,121.14) ;
\draw [shift={(214.2,124.14)}, rotate = 270] [fill={rgb, 255:red, 0; green, 0; blue, 0 }  ][line width=0.08]  [draw opacity=0] (5.36,-2.57) -- (0,0) -- (5.36,2.57) -- cycle    ;
\draw    (286.9,69.06) -- (286.9,123.14) ;
\draw [shift={(286.9,126.14)}, rotate = 270] [fill={rgb, 255:red, 0; green, 0; blue, 0 }  ][line width=0.08]  [draw opacity=0] (5.36,-2.57) -- (0,0) -- (5.36,2.57) -- cycle    ;
\draw [color={rgb, 255:red, 199; green, 2; blue, 202 }  ,draw opacity=1 ]   (250.24,136.28) -- (250.24,167.48) -- (260.87,167.48) -- (268.74,167.48) ;
\draw [shift={(271.74,167.48)}, rotate = 180] [fill={rgb, 255:red, 199; green, 2; blue, 202 }  ,fill opacity=1 ][line width=0.08]  [draw opacity=0] (5.36,-2.57) -- (0,0) -- (5.36,2.57) -- cycle    ;
\draw    (137.25,91.53) -- (137.25,212.47) ;
\draw  [fill={rgb, 255:red, 152; green, 249; blue, 255 }  ,fill opacity=1 ] (152.26,85.76) .. controls (152.26,83.62) and (153.99,81.89) .. (156.12,81.89) -- (174.34,81.89) .. controls (176.47,81.89) and (178.2,83.62) .. (178.2,85.76) -- (178.2,97.36) .. controls (178.2,99.49) and (176.47,101.22) .. (174.34,101.22) -- (156.12,101.22) .. controls (153.99,101.22) and (152.26,99.49) .. (152.26,97.36) -- cycle ;

\draw  [fill={rgb, 255:red, 249; green, 148; blue, 148 }  ,fill opacity=1 ] (190.64,86.13) .. controls (190.64,83.79) and (192.54,81.89) .. (194.88,81.89) -- (231.71,81.89) .. controls (234.05,81.89) and (235.95,83.79) .. (235.95,86.13) -- (235.95,98.85) .. controls (235.95,101.2) and (234.05,103.1) .. (231.71,103.1) -- (194.88,103.1) .. controls (192.54,103.1) and (190.64,101.2) .. (190.64,98.85) -- cycle ;
\draw  [color={rgb, 255:red, 0; green, 0; blue, 0 }  ,draw opacity=1 ][fill={rgb, 255:red, 244; green, 137; blue, 246 }  ,fill opacity=1 ] (195.46,129.03) .. controls (195.46,126.52) and (197.5,124.48) .. (200.01,124.48) -- (227.72,124.48) .. controls (230.24,124.48) and (232.28,126.52) .. (232.28,129.03) -- (232.28,142.69) .. controls (232.28,145.21) and (230.24,147.25) .. (227.72,147.25) -- (200.01,147.25) .. controls (197.5,147.25) and (195.46,145.21) .. (195.46,142.69) -- cycle ;

\draw  [fill={rgb, 255:red, 244; green, 137; blue, 246 }  ,fill opacity=1 ] (271.4,131.14) .. controls (271.4,128.84) and (273.27,126.97) .. (275.58,126.97) -- (299.06,126.97) .. controls (301.36,126.97) and (303.23,128.84) .. (303.23,131.14) -- (303.23,143.67) .. controls (303.23,145.98) and (301.36,147.85) .. (299.06,147.85) -- (275.58,147.85) .. controls (273.27,147.85) and (271.4,145.98) .. (271.4,143.67) -- cycle ;

\draw [color={rgb, 255:red, 199; green, 2; blue, 202 }  ,draw opacity=1 ]   (232.53,135.88) -- (268.87,135.88) ;
\draw [shift={(271.87,135.88)}, rotate = 180] [fill={rgb, 255:red, 199; green, 2; blue, 202 }  ,fill opacity=1 ][line width=0.08]  [draw opacity=0] (5.36,-2.57) -- (0,0) -- (5.36,2.57) -- cycle    ;
\draw  [fill={rgb, 255:red, 131; green, 131; blue, 131 }  ,fill opacity=1 ] (321.17,137.87) .. controls (321.17,134.17) and (324.17,131.17) .. (327.87,131.17) .. controls (331.57,131.17) and (334.57,134.17) .. (334.57,137.87) .. controls (334.57,141.57) and (331.57,144.57) .. (327.87,144.57) .. controls (324.17,144.57) and (321.17,141.57) .. (321.17,137.87) -- cycle ; \draw   (321.17,137.87) -- (334.57,137.87) ; \draw   (327.87,131.17) -- (327.87,144.57) ;
\draw  [fill={rgb, 255:red, 244; green, 137; blue, 246 }  ,fill opacity=1 ] (271.4,166.54) .. controls (271.4,164.24) and (273.27,162.37) .. (275.58,162.37) -- (299.06,162.37) .. controls (301.36,162.37) and (303.23,164.24) .. (303.23,166.54) -- (303.23,179.07) .. controls (303.23,181.38) and (301.36,183.25) .. (299.06,183.25) -- (275.58,183.25) .. controls (273.27,183.25) and (271.4,181.38) .. (271.4,179.07) -- cycle ;

\draw  [fill={rgb, 255:red, 244; green, 137; blue, 246 }  ,fill opacity=1 ] (271.4,201.94) .. controls (271.4,199.64) and (273.27,197.77) .. (275.58,197.77) -- (299.06,197.77) .. controls (301.36,197.77) and (303.23,199.64) .. (303.23,201.94) -- (303.23,214.47) .. controls (303.23,216.78) and (301.36,218.65) .. (299.06,218.65) -- (275.58,218.65) .. controls (273.27,218.65) and (271.4,216.78) .. (271.4,214.47) -- cycle ;

\draw [color={rgb, 255:red, 199; green, 2; blue, 202 }  ,draw opacity=1 ]   (250.24,167.48) -- (250.24,206.5) -- (268.2,206.5) ;
\draw [shift={(271.2,206.5)}, rotate = 180] [fill={rgb, 255:red, 199; green, 2; blue, 202 }  ,fill opacity=1 ][line width=0.08]  [draw opacity=0] (5.36,-2.57) -- (0,0) -- (5.36,2.57) -- cycle    ;
\draw  [fill={rgb, 255:red, 152; green, 249; blue, 255 }  ,fill opacity=1 ] (152.26,167.29) .. controls (152.26,165.15) and (153.99,163.42) .. (156.12,163.42) -- (174.34,163.42) .. controls (176.47,163.42) and (178.2,165.15) .. (178.2,167.29) -- (178.2,178.89) .. controls (178.2,181.02) and (176.47,182.76) .. (174.34,182.76) -- (156.12,182.76) .. controls (153.99,182.76) and (152.26,181.02) .. (152.26,178.89) -- cycle ;

\draw    (303.1,137.87) -- (318.17,137.87) ;
\draw [shift={(321.17,137.87)}, rotate = 180] [fill={rgb, 255:red, 0; green, 0; blue, 0 }  ][line width=0.08]  [draw opacity=0] (5.36,-2.57) -- (0,0) -- (5.36,2.57) -- cycle    ;
\draw    (304.37,175.6) -- (327.87,175.6) -- (327.87,147.57) ;
\draw [shift={(327.87,144.57)}, rotate = 90] [fill={rgb, 255:red, 0; green, 0; blue, 0 }  ][line width=0.08]  [draw opacity=0] (5.36,-2.57) -- (0,0) -- (5.36,2.57) -- cycle    ;
\draw    (334.57,137.87) -- (368.57,137.87) ;
\draw [shift={(371.57,137.87)}, rotate = 180] [fill={rgb, 255:red, 0; green, 0; blue, 0 }  ][line width=0.08]  [draw opacity=0] (5.36,-2.57) -- (0,0) -- (5.36,2.57) -- cycle    ;
\draw  [fill={rgb, 255:red, 152; green, 249; blue, 255 }  ,fill opacity=1 ] (325.82,186.36) .. controls (325.82,184.22) and (327.55,182.49) .. (329.69,182.49) -- (347.9,182.49) .. controls (350.04,182.49) and (351.77,184.22) .. (351.77,186.36) -- (351.77,197.96) .. controls (351.77,200.09) and (350.04,201.82) .. (347.9,201.82) -- (329.69,201.82) .. controls (327.55,201.82) and (325.82,200.09) .. (325.82,197.96) -- cycle ;

\draw  [fill={rgb, 255:red, 152; green, 249; blue, 255 }  ,fill opacity=1 ] (326.02,215.96) .. controls (326.02,213.82) and (327.75,212.09) .. (329.89,212.09) -- (348.1,212.09) .. controls (350.24,212.09) and (351.97,213.82) .. (351.97,215.96) -- (351.97,227.56) .. controls (351.97,229.69) and (350.24,231.42) .. (348.1,231.42) -- (329.89,231.42) .. controls (327.75,231.42) and (326.02,229.69) .. (326.02,227.56) -- cycle ;

\draw    (313.17,208.38) -- (313.17,222.78) -- (323.57,222.78) ;
\draw [shift={(326.57,222.78)}, rotate = 180] [fill={rgb, 255:red, 0; green, 0; blue, 0 }  ][line width=0.08]  [draw opacity=0] (5.36,-2.57) -- (0,0) -- (5.36,2.57) -- cycle    ;
\draw    (302.57,208.38) -- (313.17,208.38) -- (313.17,192.38) -- (323.57,192.38) ;
\draw [shift={(326.57,192.38)}, rotate = 180] [fill={rgb, 255:red, 0; green, 0; blue, 0 }  ][line width=0.08]  [draw opacity=0] (5.36,-2.57) -- (0,0) -- (5.36,2.57) -- cycle    ;
\draw    (352.37,191.98) -- (369.97,191.98) ;
\draw [shift={(372.97,191.98)}, rotate = 180] [fill={rgb, 255:red, 0; green, 0; blue, 0 }  ][line width=0.08]  [draw opacity=0] (5.36,-2.57) -- (0,0) -- (5.36,2.57) -- cycle    ;
\draw    (352.37,221.58) -- (369.17,221.58) ;
\draw [shift={(372.17,221.58)}, rotate = 180] [fill={rgb, 255:red, 0; green, 0; blue, 0 }  ][line width=0.08]  [draw opacity=0] (5.36,-2.57) -- (0,0) -- (5.36,2.57) -- cycle    ;
\draw    (177.58,91.71) -- (187.64,91.71) ;
\draw [shift={(190.64,91.71)}, rotate = 180] [fill={rgb, 255:red, 0; green, 0; blue, 0 }  ][line width=0.08]  [draw opacity=0] (5.36,-2.57) -- (0,0) -- (5.36,2.57) -- cycle    ;
\draw    (136.58,91.53) -- (149.2,91.53) ;
\draw [shift={(152.2,91.53)}, rotate = 180] [fill={rgb, 255:red, 0; green, 0; blue, 0 }  ][line width=0.08]  [draw opacity=0] (5.36,-2.57) -- (0,0) -- (5.36,2.57) -- cycle    ;
\draw    (178.25,173.71) -- (268.2,173.71) ;
\draw [shift={(271.2,173.71)}, rotate = 180] [fill={rgb, 255:red, 0; green, 0; blue, 0 }  ][line width=0.08]  [draw opacity=0] (5.36,-2.57) -- (0,0) -- (5.36,2.57) -- cycle    ;
\draw    (137.25,212.47) -- (268.4,212.47) ;
\draw [shift={(271.4,212.47)}, rotate = 180] [fill={rgb, 255:red, 0; green, 0; blue, 0 }  ][line width=0.08]  [draw opacity=0] (5.36,-2.57) -- (0,0) -- (5.36,2.57) -- cycle    ;
\draw    (104.26,173.2) -- (137.91,173.2) ;

\draw (85,171) node [anchor=north west][inner sep=0.75pt]  [font=\footnotesize]  {$\mathbf{e}_{1:n}$};
\draw (187.7,109) node [anchor=north west][inner sep=0.75pt]  [font=\footnotesize]  {$m_{1:n}$};
\draw (203.5,174.67) node [anchor=north west][inner sep=0.75pt]  [font=\footnotesize]  {$\mathbf{\delta }_{1:n}$};
\draw (276.9,56) node [anchor=north west][inner sep=0.75pt]  [font=\footnotesize]  {$\mathbf{z}_{1:n}$};
\draw (373.2,128.03) node [anchor=north west][inner sep=0.75pt]  [font=\footnotesize]  {$\tilde{\mathbf{z}}_{1:k}$};
\draw (373.7,184.14) node [anchor=north west][inner sep=0.75pt]  [font=\footnotesize]  {$\mathbf{o}_{1:k}$};
\draw (373.2,213.24) node [anchor=north west][inner sep=0.75pt]  [font=\footnotesize]  {$\mathbf{q}_{1:k}^{d}$};
\draw (154.51,86.89) node [anchor=north west][inner sep=0.75pt]  [font=\scriptsize] [align=left] {FFN};
\draw (200.33,130.21) node [anchor=north west][inner sep=0.75pt]  [font=\scriptsize] [align=left] {Top-k};
\draw (273.7,132.36) node [anchor=north west][inner sep=0.75pt]  [font=\scriptsize] [align=left] {Index};
\draw (273.7,167.76) node [anchor=north west][inner sep=0.75pt]  [font=\scriptsize] [align=left] {Index};
\draw (273.7,203.16) node [anchor=north west][inner sep=0.75pt]  [font=\scriptsize] [align=left] {Index};
\draw (241.83,118.47) node [anchor=north west][inner sep=0.75pt]  [font=\footnotesize,color={rgb, 255:red, 199; green, 2; blue, 202 }  ,opacity=1 ]  {$i_{1:k}$};
\draw (154.51,168.42) node [anchor=north west][inner sep=0.75pt]  [font=\scriptsize] [align=left] {FFN};
\draw (328.07,187.49) node [anchor=north west][inner sep=0.75pt]  [font=\scriptsize] [align=left] {FFN};
\draw (328.27,217.09) node [anchor=north west][inner sep=0.75pt]  [font=\scriptsize] [align=left] {FFN};
\draw (194.2,87.13) node [anchor=north west][inner sep=0.75pt]  [font=\scriptsize] [align=left] {Sigmoid};

\end{tikzpicture}

    \caption{MT3v2's selection mechanism: Embeddings $\mathbf e_{1:n}$ are fed to an FFN and then a sigmoid layer, producing scores $m_{1:n}$. The embeddings of the measurements with the top-k scores are fed to two FFNs, producing the object queries $\mathbf o_{1:k}$ and their corresponding positional encodings $\mathbf q_{1:k}^d$.}
    \label{fig:selection_mechanism}
    \vspace{-2mm}
\end{figure}

The selection mechanism of MT3v2, illustrated in detail in Fig.\,\ref{fig:selection_mechanism}, is in charge of producing the initial estimates for iterative refinement $\tilde{\mathbf z}_{1:k}$ (see Section \ref{subsec:iterative_refinement}), and the object queries $\mathbf o_{1:k}$ along with their positional encodings $\mathbf q_{1:k}^d$, similar to the two-stage encoder proposed in \cite{deformable-DETR}. It does so by learning to look for the measurements among $\mathbf z_{1:n}$ that are the best candidates to be used as starting points for the decoder (i.e., measurements that are likely to be close to the state estimates the decoder is in charge of producing for that specific sequence $\mathbf z_{1:n}$), and basing its outputs on them. This simplifies the decoder's task and improves performance.

First, scores $m_i\in[0, 1]$ for each of the embeddings $\mathbf e_i$ are computed according to $m_i = \textrm{Sigmoid}(\textrm{FFN}(\mathbf e_i)), i\in\mathbb N_n$.
The indices $i_{1:k}$ of the top-k scores are then computed according to $\textrm{top-k}(m_{1:n}) = [i_1, i_2, \cdots, i_k]$, where
\begin{align}
        i_j &= \arg\max_a m_a\quad\text{ s.t. } a\notin\{i_l : l<j\},
\end{align}
for $j\in\mathbb N_k$. These indices $i_{1:k}$ are used to index the sequences $\boldsymbol \delta_{1:n}$ (predicted adjustments), $\mathbf z_{1:n},$ and $\mathbf e_{1:n}$, by applying the $\textrm{Index}$ function
\begin{equation}
    \textrm{Index}(\mathbf a_{1:n}, i_{1:k}) = [\mathbf a_{i_1}, \mathbf a_{i_2}, \cdots, \mathbf a_{i_k}]~,
\end{equation}
which we will abbreviate as $\textrm{Index}(\mathbf a_{1:n}, i_{1:k})=\mathbf a_{i_{1:k}}$. The initial estimates $\tilde{\mathbf z}_{1:k}$ are then computed by summing the top-k measurements and their corresponding predicted adjustments
\begin{equation}
        \tilde{\mathbf z}_{1:k}=\mathbf z_{i_{1:k}}+\boldsymbol \delta_{i_{1:k}}~,
\end{equation}
where $\boldsymbol \delta_i = \textrm{FFN}(\mathbf e_i)$. At the same time, the object queries and decoder positional encodings are computed by feeding the top-k embeddings to separate FFN layers:
\begin{align}
    \mathbf o_{1:k}&=\textrm{FFN}(\mathbf e_{i_{1:k}}),\quad \mathbf q_{1:k}^d=\textrm{FFN}(\mathbf e_{i_{1:k}}).
\end{align}

\subsection{Iterative Refinement}
\label{subsec:iterative_refinement}
To further improve the performance of MT3v2, we adopt the idea of iterative refinement \cite{iterative_refinement_1, iterative_refinement_3, deformable-DETR} in the decoder. As stated previously, the decoder outputs the sequence $\mathbf y_{1:k}$ which represents the $k$ components of an MB density, where each $\mathbf y_i = (\boldsymbol \mu_i, \boldsymbol \Sigma_i, p_i)$ contains respectively the mean, covariance, and existence probability for that Bernoulli.  Instead of directly computing the sequence of predicted state means $\boldsymbol\mu_{1:k}$ from the output of the decoder's last layer (e.g., $\boldsymbol\mu_{1:k} = f(\mathbf o_{1:k}^{(M)})$, with some learnable $f$), we start with the initial state estimates $\tilde{\mathbf z}_{1:k}$ computed by the selection mechanism, and each decoder layer $l\in\mathbb N_M$ generates adjustments $\boldsymbol\Delta_{1:k}^l$ to it. Summing all adjustments to the initial estimates then yields the output $\boldsymbol \mu_{1:k}$ for the decoder.

Concretely, the initial estimates $\tilde{\mathbf z}_i$ are first transformed from measurement space 
to state-space, and are then denoted by $\boldsymbol\mu_i^0$. Then, the output $\mathbf o_{1:k}^{(l)}$ of each decoder layer $l$ is fed to an FFN (each layer has an FFN with separate parameters), which then produces adjustments $\boldsymbol\Delta_{1:k}^l$ in the state space. New adjustments are added to the previous estimate at each decoder layer, resulting in predicted state means $\boldsymbol\mu_{1:k}^l$ for each layer $l$, where
\begin{equation}
    \label{eq:iterative_refinement}
    \boldsymbol\mu_{i}^l = \boldsymbol\mu_i^0 + \sum_{l=1}^M\boldsymbol\Delta_i^l, \quad l\in\mathbb N_M~,
\end{equation}
Covariances and existence probabilities are not iteratively refined, and are directly computed at each layer as 
\begin{align}
    \boldsymbol \Sigma_{1:k}^l&=\text{Diag}\Big(\text{Softplus}\big(\textrm{FFN}(\mathbf o_{1:k}^{(l)})\big)\Big),
    \\
    p_{1:k}^l&=\text{Sigmoid}\big(\textrm{FFN}(\mathbf o_{1:k}^{(l)})\big),
\end{align}
where $\text{Softplus}(\cdot)$ is applied element-wise as
\begin{equation}
    \text{Softplus}(x) = \log(1+e^x),
\end{equation}
and $\text{Diag}:\mathbb{R}^n\to \mathbb{R}^{n\times n}$, also applied element-wise, is a function that constructs a diagonal matrix from its input. 
Predicting diagonal covariance matrices $\boldsymbol \Sigma_i^l$ may impact performance, but improves training time considerably (avoids the need to invert a full positive definite matrix when computing the log-likelihood of the state density, see Sec. \ref{subsec:loss}). Putting these together, each decoder layer produces an MB density $\mathbf y_{1:k}^l=(\boldsymbol\mu_{1:k}^l, \boldsymbol\Sigma_{1:k}^l, p_{1:k}^l)$. The final output of MT3v2 is then the output at the last decoder layer, i.e., $\mathbf y_{1:k}=\mathbf y_{1:k}^M$, whereas the other outputs $\mathbf y_{1:k}^l, l\in\mathbb N_{M-1}$ are used only during training (see Section \ref{subsec:loss}).

\subsection{Loss}
\label{subsec:loss}
We train MT3v2 using an approximation of the expected sum of the negative log-likelihood (NLL) of the $M$ MBs $\mathbf y_{1:k}^{l}, l\in\mathbb N_M$, evaluated at the ground-truth target states \cite{nll_letter}. Training all the intermediate decoder block outputs instead of just the final predictions $\mathbf y_{1:k}$ from the final layer is shown to accelerate learning for deep Transformer decoder architectures and improve final performance \cite{auxiliary_decoding_losses, DETR}, and is confirmed by our studies (experiments without this change often obtained considerably worse performance). We sample a measurement sequence $\mathbf z_{1:n}$, and corresponding ground-truth targets $\mathbf x_{1:m}$ using the available models of the environment, where $\mathbf x_i\in\mathbb R^{d_x},i\in\mathbb N_m$ are the states for the $m$ objects which are alive at the last time-step. Then the measurements are fed to MT3v2, which generates predictions $\mathbf y_{1:k}^l, l\in\mathbb N_M$ (one for each decoder layer, see Section \ref{subsec:iterative_refinement}). The loss for this sample is then expressed as
\begin{equation}
    \label{eq:nll_loss}
    \mathcal L(\mathbf x_{1:m}, \mathbf y_{1:k}^1, \cdots, \mathbf y_{1:k}^M) = -\sum_{l=1}^M \log f^l(\mathbf x_{1:m}),
\end{equation}
where $f^l(\mathbf x_{1:m})$ is the MB density specified by $\mathbf y_{1:k}^l$, evaluated at $\mathbf x_{1:m}$.

Computing $f^l(\mathbf x_{1:m})$ directly is computationally intractable, since all possible associations between the MB components and the ground-truth object states must be accounted for, making the number of terms in this expression grow super-exponentially on $k$ and $m$ \cite{nll_letter, crouse2016implementing}. However, in most cases all but one of the possible associations between Bernoullis and targets have negligible contribution, so we can approximate this NLL with only the contribution from the most likely association. To do so, we append `$\varnothing$' elements to the sequence $\mathbf x_{1:m}$ resulting in a new sequence $\tilde{\mathbf x}_{1:k}$ with the same number of elements\footnote{We choose a value of $k$ which is large in comparison to the generative model, and enforce $m\leq k$ by not adding more than $k$ objects to any sample. This restriction is only enforced during training; during evaluation/inference this loss needs not be computed.} as each $\mathbf y_{1:k}^l$, and approximate the NLL as
\begin{align}
    \log f^l(\mathbf x_{1:m}) & = \sum_{i=1}^k \log \sum_{\sigma} f_i^l(\tilde{\mathbf x}_{\sigma(i)})\\
    & \approx \sum_{i=1}^k \log f_i^l(\tilde{\mathbf x}_{\sigma^l(i)}),
\end{align}
where $\sigma$ is a permutation function, $\sigma: \mathbb N_k \rightarrow \mathbb N_k ~|~ \sigma(i) = \sigma(j) \Rightarrow i = j,$ corresponding to one possible association between MB components and ground-truth object states, and $f_i^l(\mathbf x_{\sigma(i)})$ is the Bernoulli density specified by $\mathbf y_i^l$ evaluated at the $\sigma(i)$-th element of $\tilde{\mathbf x}_{1:k}$, such that
\begin{equation}
\label{eq:log_bernoulli_density}
    -\log f_i^l(\tilde{\mathbf x}_j) = 
    \begin{cases}
        \log p_i^l+\log\mathcal N(\tilde{\mathbf x}_j; \boldsymbol\mu_i^l, \boldsymbol\Sigma_i^l) & \text{if } \tilde{\mathbf x}_j\neq\varnothing
        \\
        \log(1-p_i^l) & \text{otherwise.}
    \end{cases}    
\end{equation}
Finally, $\sigma^l$ 
corresponds to the most likely association between objects and Bernoulli components predicted at the decoder layer $l$. Computing $\sigma^l$ directly as described in \cite{nll_letter} resulted in unstable learning, and we instead approximate it similarly to \cite{DETR}, as 
\begin{align}
    \sigma^l &= \arg\min_\sigma \sum_{i=1}^k \mathcal L_\text{match}(\mathbf y_i^l, \tilde{\mathbf x}_{\sigma(i)}),
\end{align}
where
\begin{align}
    \mathcal L_\text{match}(\mathbf y_i^l, \tilde{\mathbf x}_{\sigma(i)}) &= 
    \begin{cases}
        0 & \text{ if } \tilde{\mathbf x}_{\sigma(i)}^l=\varnothing 
        \\ 
        \Vert \boldsymbol\mu_i^l - \tilde{\mathbf x}_{\sigma(i)}\Vert - \log p_i^l & \text{ otherwise, } 
    \end{cases}
\end{align}
which can be solved efficiently using the Hungarian algorithm \cite{hungarian-method}.

\vspace{-1mm}
\subsection{Contrastive Auxiliary Learning}
Another improvement we add to the training process is an auxiliary task of trying to predict which of the measurements in $\mathbf z_{1:n}$ came from which objects (and which are clutter). Adding simpler auxiliary tasks often improves the initial part of the training process (when the main task is still too hard to solve, and might not provide much gradient information) and the generalization performance of the final model \cite{multi_task_learning}. 

To implement this, we use an idea inspired by Supervised Contrastive Learning \cite{SCL}, where the model is trained to generate similar predictions for samples of the same classes, but dissimilar to samples of other classes. During the data generation we annotate each measurement $\mathbf z_i, i\in\mathbb N_n$, with an integer $b_i$ encoding from which object it came from, $-1$ if it is clutter
. Let $\mathbb P_i$ be the set of indices of measurements that came from the same object as the measurement $\mathbf z_i$, $\mathbb P_i = \{j \in \mathbb N_n ~|~ j\neq i~,~b_i = b_j\},$ the auxiliary loss $\mathcal L_c$ is then defined similarly to \cite{SCL}, but using the object identifiers $b_{1:n}$ as the labels for the contrastive learning of the encoder embeddings:
\begin{align}
    \label{eq:contrastive_loss}
    \mathcal L_c(\mathbf e_{1:n}, b_{1:n}) &=\beta\sum_{i=1}^n\frac{-1}{|\mathbb P_i|}\sum_{i^+\in\mathbb P_i}\log \frac{e^{\mathbf u_i^\top \mathbf u_{i^+}}}{\sum\limits_{j\in \mathbb N_n\setminus i}e^{\mathbf u_i^\top \mathbf u_j}}
    \\
    \mathbf u_{1:n} &= \frac{\textrm{FFN}(\mathbf e_{1:n})}{\Vert\textrm{FFN}(\mathbf e_{1:n})\Vert}_2.
\end{align}
where $\beta\geq 0$ is a hyperparameter controlling the trade-off between the auxiliary task and the main task. This loss can be intuitively understood as encouraging the processed embeddings $\mathbf u_i$ and $\mathbf u_j$ from different measurements $\mathbf z_i, \mathbf z_j$ to be similar if $b_i=b_j$ ($\mathbf u_i^\top\mathbf u_j$ will be large) or dissimilar if $b_i\neq b_j$ ($\mathbf u_i^\top\mathbf u_j$ will be small). Training the model on the sum of this auxiliary loss and the loss defined in Section \ref{subsec:loss} accelerated learning and improved final performance of MT3v2, specially in more challenging tasks.

\subsection{Preprocessing}
\label{subsec:preprocessing}
Aside from preprocessing techniques commonly used in DL (e.g., normalizing input, normalizing output, removing the mean), we perform two additional transformations. First, in order to use self-attention layers of dimensionality higher than $\mathbb R^{d_z}$, we increase the dimension of each measurement before feeding it to the encoder through a linear transformation
\begin{equation}
    \mathbf z'_i = \mathbf{W} \mathbf z_i + \mathbf b, \quad i\in\mathbb N_n,
\end{equation}
where $\mathbf z'_i\in\mathbb R^{d'}$ is the dimensionality augmented measurement vector, $d'>d_z$ is the new dimensionality, and $\mathbf{W}\in\mathbb R^{d'\times d_z}$, $\mathbf b\in\mathbb R^{d'}$ are learnable parameters. Second, the positional encodings $\mathbf q_{1:k}^e$ added to the input of every encoder block are computed as a learnable lookup-table that depends on the relative time-step the measurement was obtained, not on its position on the sequence: $\mathbf q_i^e=f_\lambda(t_i)$, where $t_i$ is the time-step the measurement $\mathbf z_i$ was obtained, and $\lambda$ is a parameter vector trained jointly with the other parameters of the network. This allows the architecture to have direct access to the time of measurement for each $\mathbf z_i$, while at the same time sidestepping the need to learn that the position in the sequence $\mathbf z_{1:n}$ is not relevant to the task (only the corresponding time of measurement).

\section{Evaluation Setting}
\label{sec:evaluation_setting}
This section describes the setting used to evaluate the capabilities of the proposed DL tracker in model-based MOT. Specifically, we benchmark MT3v2 against two SOTA Bayesian RFS MOT filters: the PMBM filter \cite{PMBM} and the $\delta$-GLMB filter \cite{GLMB}, in a simulated scenario with synthetic radar measurements. The PMBM filter provides a closed-form solution for MOT with standard multitarget models with Poisson birth, whereas the $\delta$-GLMB filter provides a closed-form solution for MOT when the object birth model is a multi-Bernoulli (mixture). In what follows, we first describe the tasks the different algorithms were deployed on, along with their most relevant implementation details, and then we present the measures for evaluating the filtering performance.

\subsection{Task Description}
\label{subsec:task_description}
We compare the performance of the DL approach to the traditional Bayesian filters in four different tasks. task 1 is a baseline task, simpler than the other 3, where we expect traditional approaches to be a strong benchmark for the DL tracker. We then investigate the impact of increasing the complexity of the data association (task 2), increasing the non-linearity of the models (task 3), and both changes simultaneously (task 4). 

The motion model used for all four tasks is the nearly constant velocity model, defined as:
\begin{equation}
    f(\mathbf x^{t+1}|\mathbf x^t) = 
    \mathcal N\left(\mathbf x^{t+1}; \begin{bmatrix}
        \mathbf I & \mathbf I\Delta_t
        \\ 
        \mathbf 0 & \mathbf I
    \end{bmatrix} \mathbf x^t~,~\sigma_q^2
    \begin{bmatrix}
        \mathbf I\frac{\Delta_t^3}{3} & \mathbf I\frac{\Delta_t^2}{2}
        \\
        \mathbf I\frac{\Delta_t^2}{2} & \mathbf I\Delta_t
    \end{bmatrix}\right),
\end{equation}
where $\mathbf x^{t+1}, \mathbf x^t\in \mathbb R^{d_x}$, $d_x=4$ represents target position and velocity in 2D at time-steps $t+1$ and $t$ respectively, and $\Delta_t=0.1$ is the sampling period, $\sigma_q$ controls the magnitude of the process noise. The state for newborn objects is sampled from $\mathcal N(\boldsymbol\mu_b, \boldsymbol\Sigma_b)$ with
\begin{equation*}
    \boldsymbol\mu_b=\begin{bmatrix}
        7\\ 
        0\\ 
        0\\ 
        0
    \end{bmatrix},
    \quad 
    \boldsymbol\Sigma_b=\begin{bmatrix}
        10 & 0 & 0 & 0\\ 
        0 & 30 & 0 & 0\\ 
        0 & 0 & 3 & 0\\ 
        0 & 0 & 0 & 3
    \end{bmatrix},
\end{equation*}
values chosen so as to have an object birth model that covers a reasonable part of the field-of-view. The measurement model used is a non-linear Gaussian model simulating a radar system: $g(\mathbf z | \mathbf x)=\mathcal N\big(\mathbf z; H(\mathbf x), \boldsymbol\Sigma(\mathbf x)\big),$ where $H$ transforms the $xy$-position and velocity state-vector $\mathbf x$ into $(r, \dot{r}, \theta)$, respectively the range, Doppler and bearing of each target. 
For tasks 3 and 4, $\boldsymbol\Sigma(\mathbf x)$ is computed according to the approach described in \cite{Zohair_5GFIM_TWC2018}, with the hyperparameters detailed in Appendix \ref{appx:ofdm_parameters}, resulting in a realistic radar measurement model with strong non-linearities close to the edges of the FOV (measurement noise increases quickly as the objects get closer to the edges). In contrast, in tasks 1 and 2 $\boldsymbol\Sigma(\mathbf x)$ is set to the constant
\begin{equation*}
    \boldsymbol\Sigma(\mathbf x) = 
    \begin{bmatrix}
        5.62\cdot10^{-3} & 0 & 0 \\
        0 & 9.56\cdot10^{-1} & 0 \\ 
        0 & 0 & 1.00\cdot10^{-2}
    \end{bmatrix}
\end{equation*}
where the values of the diagonal were chosen to make all tasks have similar measurement noise intensity in the central region of the FOV. The field-of-view for all tasks is the volume delimited by the ranges $(0.5, 150), (0, 30), (-1.3, 1.3)$ for $r$ (m), $\dot{r}$ (m/s), and $\theta$ (radians), respectively.

All tasks use Poisson models with parameter $\lambda_0$ for the initial number of objects, and have $\tau=20$, and $p_s(\cdot)=0.95$. In order to increase the data association complexity in tasks 2 and 4, certain hyperparameters of the generative model were changed, as shown in Table \ref{tab:data_association_hyperparam}. The simultaneous increase of the number of clutter measurements, process noise, and number of objects, along with a decrease of the detection probability, causes a substantial increase in the number of probable hypothesis that conventional MOT algorithms have to keep track of, making it considerably harder for them to perform well with a feasible computational complexity.
\begin{table}[]
    \vspace{-2mm}
    \centering
    \caption{Hyperparameters changed for increasing data association complexity.  \label{tab:data_association_hyperparam}}
    \begin{tabular}{cccccc}
        \toprule
        \textbf{Task} & \textbf{$\lambda_0$} & \textbf{$p_d$} & \textbf{$\lambda_c$} & \textbf{$\sigma_q$} & \textbf{$\lambda_b$} 
        \\
        \midrule
        1 & 2 & 0.95 & $4.4\cdot10^{-3}$ & 0.2 & $1.3\cdot10^{-4}$\\
        3 & 2 & 0.95 & $4.4\cdot10^{-3}$ & 0.2 & $1.3\cdot10^{-4}$\\
        2 & 6 & 0.7 & $2.6\cdot10^{-2}$ & 0.9 & $3.5\cdot10^{-4}$\\
        4 & 6 & 0.7 & $2.6\cdot10^{-2}$ & 0.9 & $3.5\cdot10^{-4}$\\
        \bottomrule
    \end{tabular}
    \vspace{-2mm}
\end{table}

\subsection{Implementation Details}
\label{subsec:implementation_details}
For all experiments on MT3v2, the increased dimensionality of the measurements is $d'=256$, we use $N=6$ encoder blocks and $M=6$ decoder blocks, multihead self-attention layers with 8 attention heads and $k=16$ object queries, embedding layers with $d^e=256$, object queries with $d^o=256$, and the contrastive loss hyperparameter $\beta$ is set to 4.0. All FFNs in the encoder and decoder blocks have 2048 hidden units, and are trained with a dropout rate of 0.1. All FFNs in the selection mechanism have 128 hidden units, while the one used for computing $\mathbf u_i$ in the contrastive loss has 256. In order to compute $\boldsymbol \mu_i^0$ in \eqref{eq:iterative_refinement}, measurements are mapped from measurement space to state-space as
\begin{align*}
    \boldsymbol \mu_i^0 &= \big(r_i\cos(\theta_i),~ r_i\sin(\theta_i),~ 0,~ 0\big)
\end{align*}
(i.e., using $0$ as initial estimates for the velocity dimensions). The model was trained using Adam \cite{ADAM} with a batch size of 32 and initial learning rate $5\cdot 10^{-5}$, and whenever the loss did not decrease for 50k consecutive gradient steps the learning rate was reduced by a factor of 4. Training was performed on a V100 GPU for 1M gradient steps in task 1 and 2, 700k in task 3, and 600k in task 4, amounting to approximately 4 days of training for each task. 
MT3v2 was implemented in Python + PyTorch, and the code to define, train, and evaluate it is made publicly available at \url{https://github.com/JulianoLagana/MT3v2}.

We proceed to describe the implementation details of PMBM and $\delta$-GLMB. PMBM uses a Poisson birth model with Poisson intensity $\lambda_b\mathcal{N}(\boldsymbol\mu_b, \boldsymbol\Sigma_b)$, and the initial Poisson intensity for undetected objects is set to $\lambda_0\mathcal{N}(\boldsymbol\mu_b, \boldsymbol\Sigma_b)$. As for $\delta$-GLMB, the multi-Bernoulli birth model is used, which contains a single Bernoulli component with probability of existence $\lambda_b$ and state density $\mathcal{N}(\boldsymbol\mu_b, \boldsymbol\Sigma_b)$. In addition, to model undetected objects existing at time 0, the multi-Bernoulli birth model at time 1 contains $2\lambda_0$ Bernoulli components, each of which has probability of existence $0.5$ and state density $\mathcal{N}(\boldsymbol\mu_b, \boldsymbol\Sigma_b)$.

To handle the non-linearity of the measurement model, the iterated posterior linearization filter (IPLF) \cite{garcia2015posterior} is incorporated in both PMBM and $\delta$-GLMB, see, e.g., \cite{garcia2021gaussian}. The IPLF is implemented using sigma points with the fifth-order cubature rule \cite{arasaratnam2009cubature} as suggested in \cite{crouse2014basic} for radar tracking with range-bearing-Doppler measurements, and the number of iterations is 5. In IPLF, the state dependent measurement noise covariance $\Sigma(\mathbf x)$ is approximated as $\Sigma(\hat{\mathbf x})$ where $\hat{\mathbf x}$ is either the mean of the predicted state density or the mean of the state density at last iteration. 

For PMBM and $\delta$-GLMB, the unknown data associations lead to an intractably large number of terms in the posterior densities. To achieve computational tractability of both PMBM and $\delta$-GLMB, it is necessary to reduce the number of parameters used to describe the posterior densities. First, gating is used to remove unlikely measurement-to-object associations, by thresholding the squared Mahalanobis distance, where the gating size is 20. Second, we use Murty’s algorithm \cite{crouse2016implementing} to find up to 200 best global hypotheses. Third, we prune hypotheses with weight smaller than $10^{-4}$. For PMBM, we also prune Bernoulli components with probability of existence smaller than $10^{-5}$ and Gaussian components in the Poisson intensity for undetected objects with weight smaller than $10^{-5}$. 

Both PMBM and $\delta$-GLMB implementations were developed in MATLAB. PMBM's implementation is based on the code available at \url{https://github.com/Agarciafernandez/MTT/tree/master/PMBM\%20filter}, and $\delta$-GLMB's on \url{http://ba-tuong.vo-au.com/rfs_tracking_toolbox_updated.zip}.

\vspace{-2mm}
\subsection{Performance Measures}
\label{subsec:performance_measures}
To evaluate the algorithms we use two performance measures: the generalized optimal sub-pattern assignment (GOSPA) metric \cite{GOSPA}, and the negative log-likelihood of the MOT posterior (NLL) \cite{nll_letter}. The GOSPA metric is considered due to its widespread use, its computational simplicity, and for being a metric in the space of sets. The NLL performance measure is used to evaluate the algorithms further, taking into account all of the uncertainties available in the predicted MOT posterior.
We compute a Monte-carlo approximation of the expected value of each performance mesaure by generating 1k samples of measurement sequences $\mathbf z_{1:n}$ and corresponding ground-truth object states $\mathbb X^T$ from the generative model. The measurement sequences are fed to each of the tracking algorithms, producing MOT posterior densities for each sample, which are then compared to the corresponding $\mathbb X^T$'s, using each of the performance measures. 


\subsubsection{GOSPA metric}
In order to compute the GOSPA \cite{GOSPA} metric, it is necessary to extract state estimates from the predicted MOT posterior for each algorithm, generating point-wise predictions for the states of objects alive at time-step $T$: $\hat{\mathbb X}=\{\hat{\mathbf x}_1, \cdots, \hat{\mathbf x}_{|\hat{\mathbb X}|}\}$. For PMBM and $\delta$-GLMB, we first process the MOT posterior by selecting the global hypothesis with largest weight (method 1, as defined in \cite{PMBM}), generating a multi-Bernoulli distribution. Then, $\hat{\mathbb X}$ is formed by selecting the means of all Bernoulli components with existence probability greater than $p_\text{cutoff}$, where $p_\text{cutoff}$ is chosen separately for each algorithm so as to minimize its GOSPA score. The MOT posterior defined by MT3v2's output $\mathbf y_{1:k}$ is already in the form of a multi-Bernoulli density, and we also form its $\hat{\mathbb X}$ by thresholding the components based on their existence probabilities.

Given a set of state estimates $\hat{\mathbb X}$, we compute the GOSPA metric between $\hat{\mathbb X}$ and the ground-truth target states $\mathbb X^T$, with $\alpha=2$ and Euclidean distance, defined as
\begin{equation}
\label{eq:gospa}
    \begin{aligned}
        &d_p^{(c, 2)}(\mathbb{\hat X}, \mathbb X)=
        \\
        &\min_{\gamma\in\Gamma}\Bigg( {\underbrace{\sum_{(i, j)\in\gamma} \Vert\hat{\mathbf{ x}}_i^T - \mathbf x_j^T\Vert_2^p}_{\text{Localization}}} + {\underbrace{\frac{c^p}{2}(|\mathbb{\hat X}|-|\gamma|)}_\text{False detections}} + {\underbrace{\frac{c^p}{2}(|\mathbb X|-|\gamma|)}_\text{Missed objects}} \Bigg)^{\frac{1}{p}}
    \end{aligned}
\end{equation}
where the minimization is over assignment sets
between the elements of $\hat{\mathbb X}$ and $\mathbb X$, such that $\gamma\subseteq\{1,\cdots,|\mathbb{\hat X}|\}\times\{1,\cdots, |\mathbb X|\}$, while $(i,j), (i,j')\in\gamma\implies j=j'$, and $(i,j), (i',j)\in\gamma\implies i=i'$. In all our experiments we use $c=10.0$, $p=1$. 

\subsubsection{NLL performance measure}
\label{subsubsec:nll}
The NLL performance measure is computed by evaluating how well the MOT posterior explains the ground-truth data \cite{nll_letter} in terms of its negative log-likelihood:
\begin{equation}
    \textrm{NLL}(f_a, \mathbb X^T) = -\log f_a(\mathbb X^T)
\end{equation}
where $f_a$ is the MOT posterior computed by tracker $a$. As explained in \cite{nll_letter}, in order for an algorithm to obtain a good NLL score, its MOT posterior must be able to explain the set of objects $\mathbb X^T$ for all samples, including potential missed objects and false detections. Algorithms like $\delta$-GLMB are therefore not suitable for being assessed with this performance measure, since the MOT posterior produced by them is unable to explain any missed objects: $f_{\delta-\text{GLMB}}(\mathbb X^T) = 0$ whenever $|\mathbb X^T|>n$, where $n$ is the number of Bernoulli components in $\delta$-GLMB's predicted posterior, therefore resulting in an average NLL score of $\infty$ (for NLL, lower values entail better performance). For PMBM we use method 1 as described in \cite{PMBM} to extract a PMB density, which is then able to explain any number of missed objects due to its PPP component.

In order to evaluate MT3v2 with the NLL performance measure, we add a PPP component with piece-wise constant intensity function described as
\begin{equation}
    \lambda_\text{MT3v2}(\mathbf x) = 
    \begin{cases}
        \bar\lambda~, \quad \text{if }\mathbf x \text{ is inside the FOV}
        \\
        0~, \quad \text{otherwise}
    \end{cases}
\end{equation}
to its posterior, resulting in a PMB MOT density for MT3v2:
\begin{equation}
    f_\text{MT3v2}(\mathbb X^T) = \sum_{\mathbb X^D\uplus\mathbb X^U=\mathbb X^T}
    f_\text{MB}(\mathbb X^D)
    e^{-\bar\lambda}\prod_{\mathbf x\in\mathbb X^U}\lambda_\text{MT3v2}(\mathbf x)
\end{equation}
where $f_\text{MB}(\cdot)$ is the MB density with $k$ Bernoulli components described by MT3v2's output $\mathbf y_{1:k}$, and $\bar\lambda$ is tuned to minimize NLL over 1k samples from the generative model.

Lastly, directly computing the NLL for a PMB density is not computationally tractable with the exception of the simplest cases, so we approximate it using the algorithm presented in \cite{nll_letter}, resulting in the following performance measure:
\begin{align}
    \label{eq:pmb_nll_decomposition}
    &\textrm{NLL}(f, \mathbb X)\approx\min_{\gamma\in\Gamma}
    \underbrace{-\sum_{(i,j)\in\gamma}\log \big(p_ig_i(\mathbf x_j)\big)}_\text{Localization}
    \\
    &\underbrace{-\sum_{i\in\mathbb F(\gamma)}\log(1-p_i)}_\text{False detections}~
    \underbrace{+\int \lambda(y')\textrm{d}y'-\sum_{j\in\mathbb M(\gamma)}\log\lambda(\mathbf x_j)}_\text{Missed objects}~,\notag
\end{align}
where $p_i, g_i$ are the existence probabilities
\footnote{
    Method 1 of extracting state estimate presented in \cite{PMBM} often generated Bernoulli components with existence probability equal to $1$. In order to do a fair comparison with MT3v2 we cap all existence probabilities for both methods at a maximum of $0.99$ instead, which prevents PMBM from obtaining too large of a penalty for some samples with false detections.
} 
and state densities for the $i$-th Bernoulli components of the PMB density $f$, and $\lambda(\cdot)$ is its PPP intensity function. Here $\Gamma$ is the set of all possible assignment sets (as defined for GOSPA), while $\mathbb F(\gamma)=\left\{ i\in\mathbb N_m ~|~ \nexists~j:(i,j)\in\gamma \right\}$ is the set of indices of the Bernoullis not matched to any ground-truth ($m$ is the number of Bernoulli components in $f$), and $\mathbb M(\gamma)=\left\{j\in\mathbb N_{|\mathbb X|}~|~\nexists~i: (i,j)\in\gamma\right\}$ is the set of indices of ground-truths not matched to any Bernoulli component.  

\section{Results}
\label{sec:results} 

This section contains the results of the evaluations performed for assessing the tracking capabilities of the proposed deep learning tracker, and is divided into three subsections. 
First, subsection \ref{subsec:illustrative_example}, describes an illustrative example of the performance of MT3v2 in a simple tracking task, depicting the predictions generated by the algorithm in this context and validating their soundness.
Second, subsection \ref{subsec:comparison_to_mb_sota} contains the results of a thorough comparison of MT3v2 to the model-based SOTA algorithms PMBM and $\delta$-GLMB, in the four tasks described in Section \ref{subsec:task_description}.
Third, subsection \ref{subsec:complexity_evaluation} further evaluates each algorithm by investigating their running time complexity.

\subsection{Illustrative Example}
\label{subsec:illustrative_example}

    As a way to validate the soundness of the MT3v2 architecture, it is helpful to illustrate the predictions generated by it given a certain sequence of measurements. However, doing so using measurement sequences sampled from any of the four tasks described in section \ref{subsec:task_description} proved to be not optimal for this. The high measurement and motion noise in these tasks, along with a high number of clutter measurements makes it challenging to interpret measurement sequences visually. 
    
\begin{figure}
        \centering
        \includegraphics[width=\columnwidth]{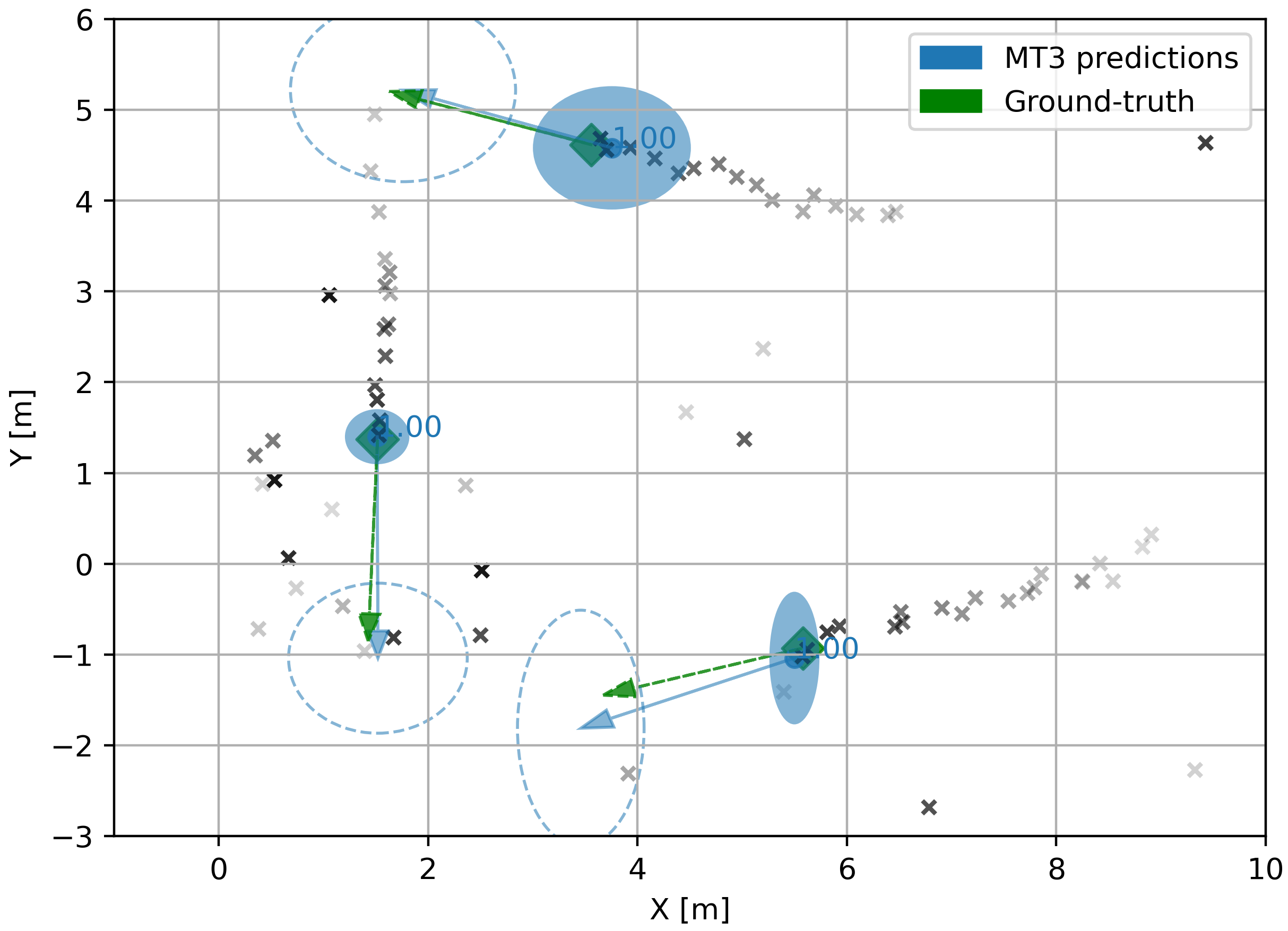}
        \caption{Illustration of MT3v2's tracking performance in a simplified setting. Black crosses illustrate the measurements available to MT3v2 (Doppler component not illustrated to avoid clutter), and their transparency is determined by their relative time of measurement: more opaque crosses correspond to more recent measurements, closer to $t=T$. Dark blue circles and arrows show MT3v2's predicted object positions and velocities. Light blue ellipses illustrate the predicted position uncertainties, and dashed ellipses the predicted velocity uncertainties. Ground-truth object positions and velocities are illustrated in green as diamonds and arrows, respectively. }
        \label{fig:mt3_tracking}
        \vspace{-2mm}
    \end{figure}

    Therefore, we resorted to creating a new, simpler task with fewer clutter measurements and lower measurement noise just for this purpose, and trained MT3v2 in it. MT3v2's tracking capabilities after training are illustrated in Fig.\,\ref{fig:mt3_tracking}, which contains the measurement sequence fed to MT3v2, along with the predictions generated for this 2D tracking task. It shows MT3v2 being able to track three objects among clutter, estimating its position and velocities and also providing sensible uncertainty predictions for these quantities.

    Evidently, although helpful as a sanity test for the approach, this type of analysis does not suffice for comparing MT3v2's performance to other approaches. Hence, we perform a thorough comparison in Section \ref{subsec:comparison_to_mb_sota} using the performance measures described in Section \ref{subsec:performance_measures} over a large number of samples instead.

\vspace{-1mm}
\subsection{Comparison to Model-Based SOTA}
\label{subsec:comparison_to_mb_sota}
    
\begin{figure}
        \centering
        \includegraphics[width=\columnwidth]{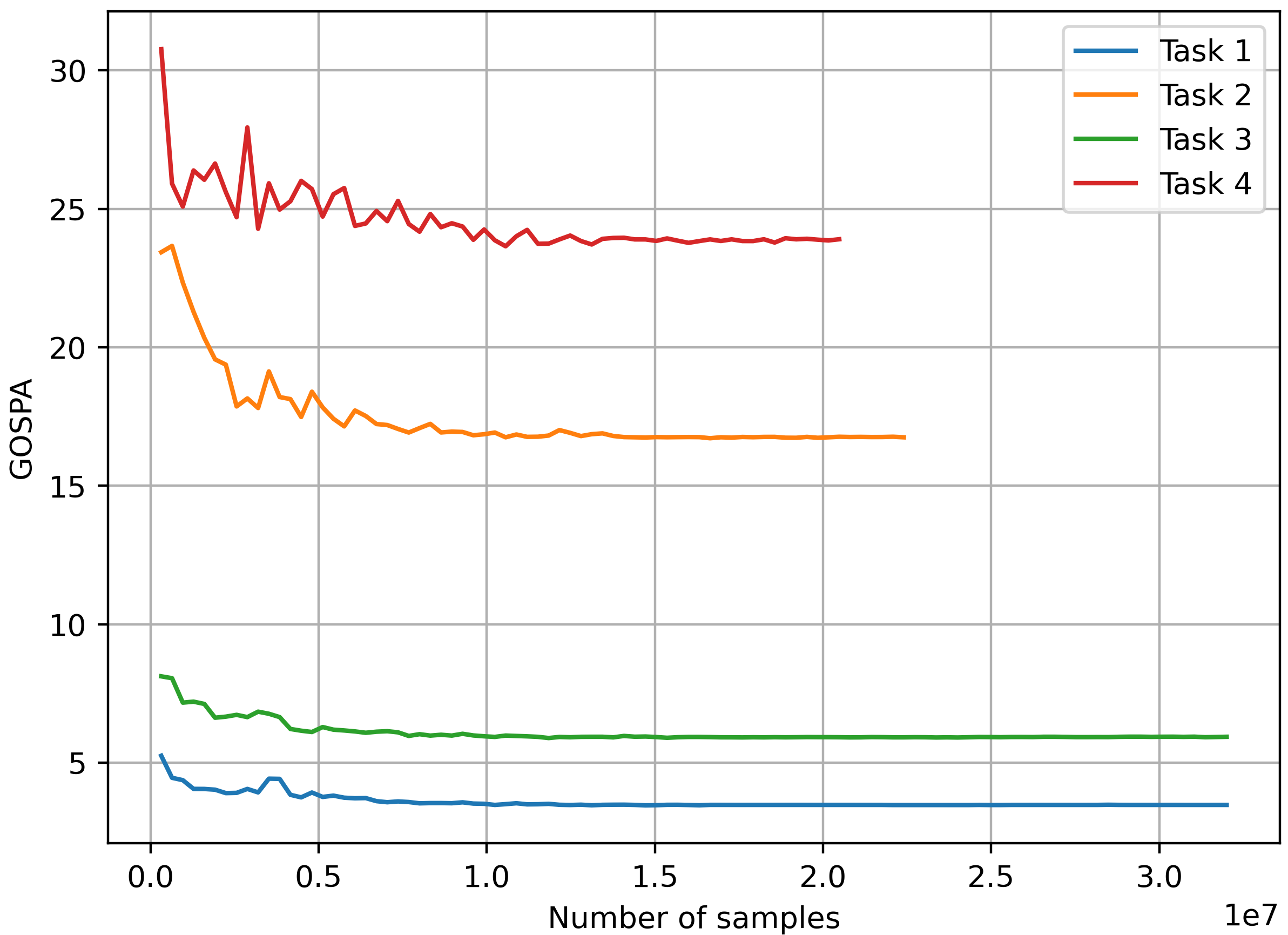}
        \caption{GOSPA scores for MT3v2 during training for each of the four tasks. Performance improves steadily with more training, and most of the gains come from the first 1M training samples. Training on tasks 2 and 4 takes longer due to the more complicated measurement model, so fewer samples were processed in the same amount of allotted training time than for the other tasks.}
        \label{fig:gospa_over_time}
    \end{figure}

    The performance of MT3v2 was compared to both SOTA Bayesian filters PMBM and $\delta$-GLMB, in the four tasks introduced in Section \ref{subsec:task_description}. For this purpose, MT3v2 was trained from scratch in each of the 4 tasks, and the GOSPA values during training for each of them are shown in figure \ref{fig:gospa_over_time}. For all tasks, performance improved steadily with more training, and most of the gains in performance are obtained within the first 1M processed samples (30k gradient steps, 1-2 days of training time). Training was continued nevertheless to ensure that the loss had plateaued. Note that the GOSPA values for tasks 2 and 4 are considerably larger than for the other tasks primarily due to a higher average number of ground-truth objects to be predicted.

    Once trained, we computed the average GOSPA and NLL scores over 1k sample for MT3v2 in each task. The resulting scores, together with those for the benchmark algorithms, are shown in Table \ref{tab:gospa_scores} and \ref{tab:nll_scores}, along with the corresponding decompositions and 95\% confidence intervals. In terms of GOSPA, we see that MT3v2 is able to match performance with the best benchmark in the simpler setting, task 1, while outperforming it in tasks 2, 3, and 4, supporting our hypothesis that DL trackers can outperform traditional model-based methods when the data association becomes more complicated and/or the models more nonlinear. In terms of NLL the conclusion is similar, with the exception of task 2, where MT3v2 has similar performance to PMBM.
    
    Additionally, for most of the tasks we notice that the localization error for MT3v2 (in both GOSPA and NLL) is higher than for the benchmarks, suggesting that the regression part of the network could be further improved. We theorize that further training plus predicting full state covariances (we only predict diagonal covariance matrices, see section \ref{subsec:iterative_refinement}) would improve this, but leave it for future work. At the same time, we note the higher GOSPA-missed and NLL-missed cost for PMBM and $\delta$-GLMB in almost all tasks, the gap to MT3v2 being the largest for task 4. As expected, the increase in the data association complexity for these tasks requires traditional approaches to aggressively prune hypothesis to remain computationally tractable, leading to more missed targets and worse performance. 
    
    \begin{table}[]
        \vspace{-2mm}
        \centering
        \caption{GOSPA scores for all tasks.  \label{tab:gospa_scores}}
        \begin{tabular}{@{}lllllll@{}}
            \toprule
            \textbf{Task} & \textbf{Algorithm} & \textbf{GOSPA} & \textbf{Localization} & \textbf{False} & \textbf{Missed} 
            \\ 
            \midrule
            \multirow{2}{*}{1} 
                & PMBM & \textbf{3.44 $\pm$ 0.18} & 2.24 & 0.12 & 1.07 
                \\
                & $\delta$-GLMB  & 3.84 $\pm$ 0.20 & 2.13 & 0.06 & 1.64 
                \\
                & MT3v2 & \textbf{3.46 $\pm$ 0.18} & 2.32 & 0.12 & 1.01 
                \\ 
            \midrule
            \multirow{2}{*}{2} 
                & PMBM & 19.03 $\pm$ 0.53 & 6.40 & 0.57 & 12.06 
                \\
                & $\delta$-GLMB  & 20.63 $\pm$ 0.61 & 5.56 & 0.23 & 14.83  
                \\
                & MT3v2 & \textbf{17.03 $\pm$ 0.46} & 8.12 & 0.96 & 7.95 
                \\ 
            \midrule
            \multirow{2}{*}{3} 
                & PMBM  & 7.27 $\pm$ 0.30 & 3.93 & 0.10 & 3.24 
                \\
                & $\delta$-GLMB  & 7.42 $\pm$ 0.30 & 3.21 & 0.70 & 3.79  
                \\
                & MT3v2 & \textbf{6.01 $\pm$ 0.27} & 3.50 & 0.21 & 2.29
                \\ 
            \midrule
            \multirow{2}{*}{4} 
                & PMBM & 26.72 $\pm$ 0.71 & 2.08 & 0.02 & 24.61 
                \\
                & $\delta$-GLMB  & 27.43 $\pm$ 0.71 & 3.26 & 0.02 & 24.14  
                \\
                & MT3v2 & \textbf{22.82 $\pm$ 0.56} & 8.88 & 0.57 & 13.37
                \\ 
            \bottomrule
        \end{tabular}
        \vspace{-2mm}
    \end{table}
    
    \begin{table}[]
        \centering
        \caption{NLL scores for all tasks.  \label{tab:nll_scores}}
        \begin{tabular}{@{}llllll@{}}
            \toprule
            \textbf{Task} & \textbf{Algorithm} & \textbf{NLL} & \textbf{Localization} & \textbf{False} & \textbf{Missed} 
            \\ 
            \midrule
            \multirow{2}{*}{1} 
                & PMBM  & \textbf{1.78 $\pm$ 0.33}  & 1.24 & 0.14 & 0.39
                \\
                & MT3v2 & 6.49 $\pm$ 0.35 & 5.76 & 0.25 & 0.48 
                \\ 
            \midrule
            \multirow{2}{*}{2} 
                & PMBM & \textbf{31.40} $\pm$ 1.00 & 23.57 & 1.47 & 6.35
                \\
                & MT3v2 & 36.00 $\pm$ 0.94 & 29.75 & 2.62 & 3.62
                \\
            \midrule
            \multirow{2}{*}{3} 
                & PMBM & 22.39 $\pm$ 1.01 & 10.85 & 2.39 & 9.16
                \\
                & MT3v2 & \textbf{12.90 $\pm$ 0.54} & 10.80 & 0.66 & 1.43
                \\
            \midrule
            \multirow{2}{*}{4} 
                & PMBM & 55.21 $\pm$ 1.49 & 25.23 & 1.18 & 28.79
                \\
                & MT3v2 & \textbf{47.67$\pm$ 1.13} & 40.25 & 3.49 & 3.94
                \\
            \bottomrule
        \end{tabular}
        \vspace{-2mm}
    \end{table}
    
\begin{figure}
        \centering
        \includegraphics[width=0.45\textwidth]{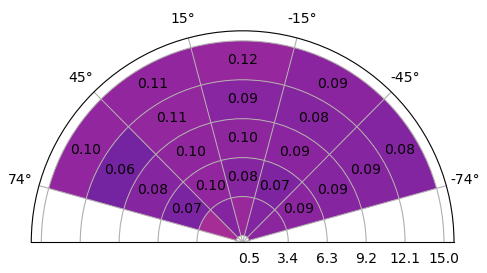}
        \caption{PMBM's missed rate per FOV sector for task 2 (very similar to MT3v2).}
        \label{fig:pmbm_missed_rate_task2}
        \vspace{-2mm}
    \end{figure}

\begin{figure}
        \vspace{-4mm}
        \centering
        \includegraphics[width=0.45\textwidth]{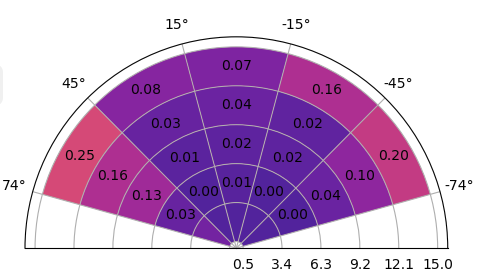}
        \caption{MT3v2's missed rate per FOV sector for task 3.}
        \label{fig:mt3v2_missed_rate_task3}
        \vspace{-2mm}
    \end{figure}

\begin{figure}
        \centering
        \includegraphics[width=0.45\textwidth]{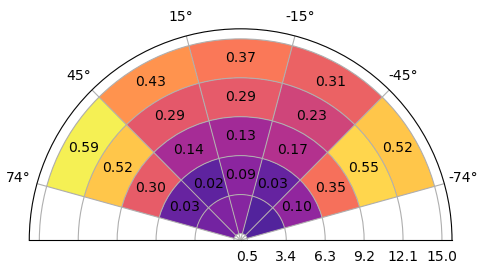}
        \caption{PMBM's missed rate per FOV sector for task 3.}
        \label{fig:pmbm_missed_rate_task3}
        \vspace{-2mm}
    \end{figure}

    To further investigate the high missed costs for the top performing benchmark PMBM, we divide the $(r, \theta)$ dimensions of the FOV into 25 different sectors, and plot PMBM's missed ratio (number of missed objects / total number of objects) in each sector for tasks 2 and 3, along with those of MT3v2, in Figures \ref{fig:pmbm_missed_rate_task2} through \ref{fig:pmbm_missed_rate_task3}. The missed rates for MT3v2 in task 2 are very similar to PMBM, so we omit it for conciseness.
    Figure \ref{fig:pmbm_missed_rate_task2} shows that in task 2 both PMBM's and MT3v2's missed ratio are reasonably constant throughout the FOV, in line with our expectations, since the data association complexity is increased across the entire FOV, with no specific region being more or less complex than others. 
    In task 3 on the other hand, Figs.\,\ref{fig:mt3v2_missed_rate_task3} and \ref{fig:pmbm_missed_rate_task3} show that PMBM's performance is considerably worse than MT3v2's for objects closer to the edges of the FOV. In these regions the measurement model becomes highly non-linear, with the state-dependent measurement noise covariance $\boldsymbol\Sigma(\mathbf x)$ (Section \ref{subsec:task_description}) increasing rapidly. We hypothesize that the Gaussian approximations in PMBM are not accurate enough in the presence of such strong non-linearities, and thus negatively impact the tracker's performance. In task 4 both of these changes (increased data association complexity and model nonlinearities) affect PMBM's performance, explaining its very high GOSPA and NLL's missed costs. In contrast, MT3v2's missed costs for all tasks are lower than for both benchmarks, specially in task 4 (3.94 vs 28.79 NLL-missed costs), suggesting that DL based trackers indeed handle these challenges in a better way than traditional model-based approaches.

\subsection{Complexity Evaluation}
\label{subsec:complexity_evaluation}

As a further comparison of the algorithms considered, this section describes the inference times\footnote{The time required to process a complete sequence of measurements $\mathbf z_{1:n}$ and generate a predicted posterior density for $\mathbb X^T$.} for MT3v2, PMBM, and $\delta$-GLMB in each of the 4 tasks from Section \ref{subsec:task_description}. MT3v2 was run on a V100 GPU, and PMBM and $\delta$-GLMB on 32 Intel Xeon Gold 6130 CPUs. The average inference times are shown in Table \ref{tab:inference_times}.
\begin{table}[]
    \centering
    \caption{Inference times for each algorithm.  
    \label{tab:inference_times}}
    \begin{tabular}{@{}llll@{}}
        \toprule
        \textbf{Task} & \textbf{Algorithm} & \textbf{Inference time (s)}
        \\ 
        \midrule
        \multirow{2}{*}{1} 
            & PMBM & 4.92 
            \\
            & $\delta$-GLMB  & 13.14
            \\
            & MT3v2 & 0.03 
            \\ 
        \midrule
        \multirow{2}{*}{2} 
            & PMBM & 126.80
            \\
            & $\delta$-GLMB  & 217.66
            \\
            & MT3v2 & 0.04 
            \\ 
        \midrule
        \multirow{2}{*}{3} 
            & PMBM  & 13.90
            \\
            & $\delta$-GLMB  & 40.40
            \\
            & MT3v2 & 0.04 
            \\ 
        \midrule
        \multirow{2}{*}{4} 
            & PMBM & 310.14 & 
            \\
            & $\delta$-GLMB  & 781.00
            \\
            & MT3v2 & 0.06  
            \\ 
        \bottomrule
    \end{tabular}
    \vspace{-2mm}
\end{table}
From the table one can see that MT3v2's inference is orders of magnitude faster than the traditional methods during inference, and can directly be used for real-time tracking in many contexts. Additionally, inference times for it scale considerably better than for the benchmarks when increasing the complexity of the task (MT3v2 is more than 5000 times faster than the benchmarks in the most complicated task), highlighting another advantage with DL-based Transformer approaches. However, we also note that this comparison between approaches is far from perfect. 

First, it is complicated to compare inference times between MT3v2 and the benchmarks, because these approaches are fundamentally different. MT3v2 is based on Transformers and deep learning, and therefore benefits greatly from parallelization and specific hardware (such as GPUs) that has been perfected over recent years to increase inference and training speed.
On the other hand, traditional Bayesian methods such as the ones we compare to rely on processing each time-step in the sequence of measurements sequentially, therefore being harder to parallelize; benefitting more from faster CPUs instead.
Second, deep learning methods require training before they can be used for inference, which as noted previously can take a considerably amount of time (4 days for each task in the case of MT3v2).
Third, MT3v2 and the Bayesian methods were run on different hardware, and implemented using different software, as described in Section \ref{subsec:implementation_details}. Our hardware choices were based on the available resources from C3SE, while our software choices mostly reflect what other open-source implementations used, rather than what would be optimal for each of the methods.

Nevertheless, the difference between inference times is so significant that we deemed worth mentioning, even if the comparison is not perfect. We expect that even in the case that considerable effort is dedicated to speeding up the benchmarks, DL-based approaches that process the measurement sequence in parallel will continue to be more efficient, specially in challenging tasks. 

\section{Conclusion}
\label{sec:conclusion}
In this paper, we propose a DL tracker based on the Transformer architecture with specific modifications to make it better suited for MOT: MT3v2. Using this tracker as a proof of concept, we compare the performance of DL versus SOTA Bayesian algorithms in the model-based multi-object tracking domain. 
Our results show that deep learning trackers can match the performance of Bayesian algorithms in simple tasks, where their performance is close to optimal, while at the same time being able to outperform them when the tracking task becomes more complicated, either due to increased complexity in the data association or stronger non-linearities in the models of the environment. This validates the applicability of deep learning to the multi-object tracking problem also in the model-based regime.

Interesting possibilities for future work are: (1) Adding more flexibility to the families of densities predicted by MT3v2 (e.g. Bernoulli components with non-diagonal covariances, more complicated state densities, normalizing flows \cite{normalizing_flows}, etc.); (2) Using better approximations to the NLL loss for training, for instance using the top-k, $k\geq 1$, associations between Bernoulli components and ground-truth \cite{nll_letter}; (3) Leveraging recent developments to the Transformer architecture (e.g. \cite{image_transformer}) for allowing MT3v2 to work efficiently with higher-dimensional measurements such as images.

\begin{appendices}

\section{OFDM parameters}
\label{appx:ofdm_parameters}
The parameters for the realistic RADAR model used as the measurement model for tasks 3 and 4 (see Section \ref{subsec:task_description}) are described in detail in Table \ref{tab:ofdm_parameters}, based on \cite{Hexa-X-D3.1}.

\begin{table}[h!]
    \vspace{-2mm}
    \centering
    \caption{OFDM parameters for realistic RADAR model}
    \label{tab:ofdm_parameters}
    \begin{tabular}{@{}ll@{}}
        \toprule
        \textbf{Parameter name}          & \textbf{Value}               \\ \midrule
        Transmission power               & $0$ dBm                    \\
        Carrier frequency                & $140$ GHz                    \\
        Noise power spectral density     & $-174$ dBm/Hz                \\
        Total bandwidth                   & $2$ GHz                     \\
        Number of subcarriers            & $1000$                       \\
        Subcarrier spacing               & $2$ MHz                      \\
        Radar cross-section              & $0.1$ m$^2$                   \\
        Receiver noise figure            & $10$ dB                      \\
        Number of receive antennas       & $20$                         \\
        Number of OFDM symbols           & $2048$                       \\
        Cyclic prefix overhead           & $7$\%           
        \\ \bottomrule
    \end{tabular}
    \vspace{-3mm}
\end{table}


\end{appendices}

\balance

\bibliographystyle{IEEEtran}
\bibliography{refs}


 






\end{document}